\documentclass[10pt,twocolumn,letterpaper]{article}

\usepackage{cvpr}              

\usepackage{url}
\usepackage{wrapfig}
\usepackage{graphicx}
\usepackage{multirow}
\usepackage{siunitx}
\usepackage{booktabs}
\usepackage{adjustbox}
\usepackage{tabularx}
\usepackage{colortbl}
\usepackage{xcolor}
\usepackage{float}
\usepackage[linesnumbered,ruled,vlined]{algorithm2e}

\definecolor{best}{rgb}{0.96, 0.57, 0.58}
\definecolor{second}{rgb}{0.98, 0.78, 0.57}
\definecolor{third}{rgb}{1.0, 1.0, 0.56}

\definecolor{cvprblue}{rgb}{0.21,0.49,0.74}
\usepackage[pagebackref,breaklinks,colorlinks,allcolors=cvprblue]{hyperref}

\def\paperID{****} 
\def\confName{CVPR}
\def\confYear{2025}

\newcommand{\cyp}[1]{\textcolor{black}{#1}}

\title{Deformable Radial Kernel Splatting}

\author{
\textbf{Yi-Hua Huang}\textsuperscript{1}
\ 
\textbf{Ming-Xian Lin}\textsuperscript{1}
\ 
\textbf{Yang-Tian Sun}\textsuperscript{1}
\ 
\textbf{Ziyi Yang}\textsuperscript{1}
\ 
\textbf{Xiaoyang Lyu}\textsuperscript{1}
\\
\textbf{Yan-Pei Cao}\textsuperscript{2$\dagger$}
\ 
\textbf{Xiaojuan Qi}\textsuperscript{1$\dagger$} \\
\textsuperscript{1} The University of Hong Kong
\quad
\textsuperscript{2} VAST
}

\begin{document}

\twocolumn[{%
\renewcommand\twocolumn[1][]{#1}%
\maketitle
\begin{center}
\vspace{-5mm}
\includegraphics[width=1\linewidth]{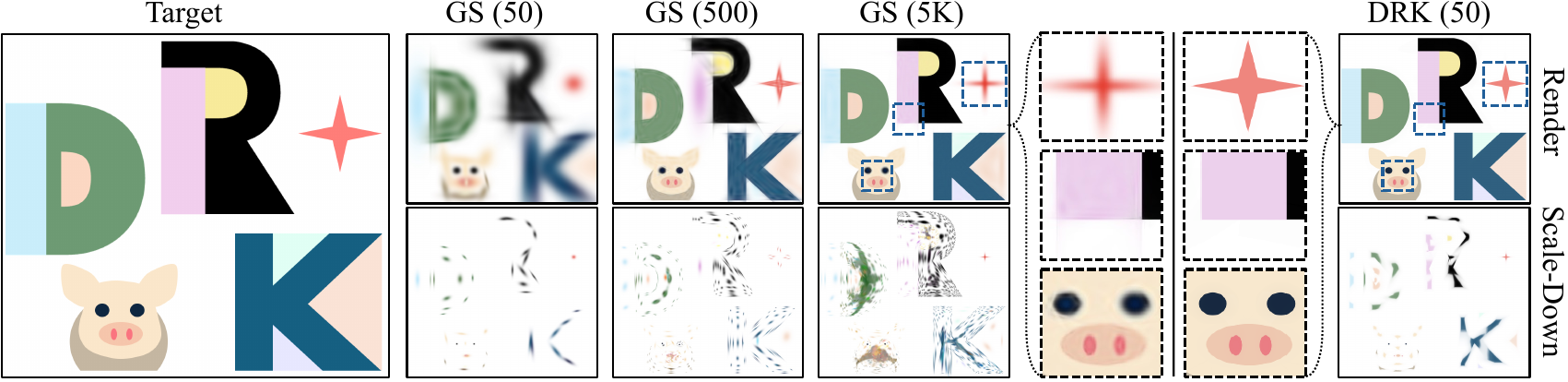}
\vspace{-6mm}
\captionof{figure}{\textbf{Gaussian Splatting vs. Our \cyp{Deformable Radial Kernel (DRK)} Splatting}: Gaussian splatting requires thousands of Gaussians to approximate detailed textures and shapes. In contrast, our kernel efficiently fits the target pattern with just 30 primitives, achieving superior results.}
\label{fig:teaser}
\end{center}
}]

\if TT\insert\footins{\noindent\footnotesize{
$\dagger$Corresponding Author\\
Project page: \url{https://yihua7.github.io/DRK-web/}.}}\fi

\begin{abstract}
Recently, Gaussian splatting has emerged as a robust technique for representing 3D scenes, enabling real-time rasterization and high-fidelity rendering. However, Gaussians' inherent radial symmetry and smoothness constraints limit their ability to represent complex shapes, often requiring thousands of primitives to approximate detailed geometry. We introduce Deformable Radial Kernel (DRK), which extends Gaussian splatting into a more general and flexible framework. Through learnable radial bases with adjustable angles and scales, DRK efficiently models diverse shape primitives while enabling precise control over edge sharpness and boundary curvature. iven DRK's planar nature, we further develop accurate ray-primitive intersection computation for depth sorting and introduce efficient kernel culling strategies for improved rasterization efficiency. Extensive experiments demonstrate that DRK outperforms existing methods in both representation efficiency and rendering quality, achieving state-of-the-art performance while dramatically reducing primitive count.
\end{abstract}
    
\section{Introduction}

3D Gaussian Splatting~\cite{kerbl2023d} (3D-GS) has emerged as a leading 3D representation method thanks to its rapid rasterization and superior rendering quality. Its key advantages-- explicit point-based representation, flexibility for manipulation, and MLP-free rendering-- have positioned it as a dominant approach in novel view synthesis.
However, this success raises an intriguing question: \emph{Is the Gaussian kernel truly optimal for representing 3D scenes?} 

Natural scenes often consist of primitives with diverse shapes (see Fig.~\ref{fig:demo}), such as rectangles, triangles, and ellipses, which cannot be fully captured by Gaussians alone. Moreover, the inherently smooth nature of Gaussian kernels makes them less effective at representing sharp transitions. As a result, densely packed or numerous Gaussians are frequently required to accurately model scene primitives with non-Gaussian geometries or sharp boundaries (see Fig.~\ref{fig:teaser}), leading to inefficiencies in both computation and memory.  

Recent works have explored modifications to the Gaussian kernel to better handle discontinuities. For example, GES~\cite{hamdi2024ges} adaptively adjusts the exponent values to control the sharpness of Gaussians, while still maintaining rotational symmetry, which limits its representational flexibility. DisC-GS~\cite{qu2024disc} and 3D-HGS~\cite{li20243d} introduce curve- and half-cutting techniques to the Gaussian kernel, respectively, improving its ability to capture discontinuous transitions. However, these cutting-based approaches remain fundamentally constrained by the inherent smoothness of the Gaussian kernel.
Another line of research~\cite{guedon2024sugar,huang20242d,gao2023relightable,dai2024high} leverages 2D Gaussians to represent 3D scenes. 
While this approach offers better surface reconstruction compared to 3DGS, it is still constrained by the inherent limitations of the Gaussian framework, which restricts the flexibility of the representation.
Notably, the concurrent work 3D-CS~\cite{Held20243DConvex} employs novel 3D smooth convex shapes as primitives for geometry modeling with satisfactory results.

\cyp{
In this work, our objective is to design new shape primitives that can adapt to complex 3D scenes. Motivated by the success of 2D Gaussian Splatting (2D-GS), we focus on developing adaptive 2D planar kernels. Specifically, we introduce the 2D Deformable Radial Kernel (\emph{DRK}), a flexible planar primitive for representing 3D scenes (see Fig.~\ref{fig:kernel}). At its core, DRK extends traditional radial bases with learnable parameters that enable precise control over shape deformation through three key features: 1) multiple radial bases with learnable polar angles and scale parameters allow the kernel to adaptively stretch, rotate, and shrink, providing significantly more flexibility than fixed Gaussian functions; 2) a hybrid distance metric combining $L1$ and $L2$ norms with learnable weights enables DRK to naturally represent both curved surfaces and sharp geometric features; and 3) an adaptive piecewise linear remapping function provides fine-grained control over value distributions, particularly enhancing the representation of sharp boundaries. Despite its enhanced expressiveness, DRK maintains a simple parametric form that is easy to optimize and efficient to render. The DRK determines the location and orientation of the tangent plane through its center and rotation parameters, building upon these radial bases as its structural foundation to conform to target shapes, bypassing the geometric and smoothness constraints typically imposed by Gaussian functions. As shown in Fig.~\ref{fig:demo}, a single DRK primitive can flexibly model a wide range of shape primitives, whereas many Gaussians are often required to represent similar shapes. The formulation is also backwards-compatible -- DRK naturally reduces to a Gaussian kernel when using two orthogonal major bases with different decay rates, while achieving superior rendering quality with comparable computational efficiency.
}

We experimentally validate \textit{DRK} on our proposed dataset (DiverseScene) covering various scenarios, including rich textures, fine geometry, specular effects, and large spatial scale. We also evaluate \textit{DRK} on unbounded Mip-NeRF 360~\cite{barron2022mip} datasets. Experimental results demonstrate our method has advantages in preserving visual details more effectively and efficiently.

Overall, our work makes the following contributions:
\begin{itemize}
    \item \cyp{We introduce Deformable Radial Kernel (\emph{DRK}), a novel primitive that generalizes conventional Gaussian splatting. DRK enables flexible shape modeling through learnable radial bases, hybrid distance metrics, and adaptive sharpness control, significantly reducing the number of primitives needed for high-quality scene representation.}
    \item \cyp{We develop a differentiable and efficient rendering pipeline for DRK, featuring polygon-based tile culling and cache-sorting strategies to enhance both rendering efficiency and view consistency.}
    \item We created DiverseScenes, a new dataset covering rich textures, fine geometry, specular effects, and large scenes for more effectively evaluating the ability to model complex scenarios of different algorithms. 
\end{itemize}

\section{Related Work}

\subsection{Novel View Synthesis}
Novel view synthesis from multi-view images represents a fundamental challenge in computer graphics and computer vision. The field has evolved through increasingly sophisticated 3D scene representations. Traditional light field approaches~\cite{gortler1996lumigraph,davis2012unstructured,levoy1996light} attempt to model the entire scene through a single 4D function of ray-slice intersections, but this implicit representation struggles with sparse inputs and complex geometry.
Advancing beyond single-function modeling, Multi-Plane-Images (MPI)~\cite{srinivasan2019pushing,tucker2020single,wizadwongsa2021nex,zhou2018stereo,mildenhall2019local} introduced a layered representation using multiple depth planes, offering better 3D structure modeling. However, its discrete depth layering limits the handling of large viewpoint changes, particularly in rotational views.
To achieve continuous depth modeling, subsequent works adopted explicit 3D proxies including meshes~\cite{waechter2014let,buehler2001unstructured,debevec1996modeling,wood2000surface}, point clouds~\cite{niklaus20193d,wiles2020synsin,meshry2019neural,aliev2020neural}, and voxels~\cite{seitz1999photorealistic,flynn2019deepview,lombardi2019neural,sitzmann2019deepvoxels}. These representations offer more flexible geometry modeling, with point-based methods further enhanced by neural networks~\cite{riegler2020free,riegler2021stable} for improved stability.

Most recently, continuous neural representations have emerged as a powerful alternative, replacing discrete geometric proxies with MLPs that directly map 3D coordinates to scene properties. NeRF~\cite{mildenhall2021nerf} pioneered this approach with radiance fields, spawning numerous extensions for faster inference~\cite{fridovich2022plenoxels,karnewar2022relu,mueller2022instant}, large-scale scenes~\cite{zhang2020nerf++,barron2022mip,tancik2022block,gao2023surfelnerf}, dynamic content~\cite{NEURIPS2022_eeb57fdf,qiao2022neuphysics,song2023nerfplayer}, reflection modeling~\cite{boss2021nerd,yang2023sire,kuang2022neroic,huang2023nerf}, and stylization~\cite{zhang2022arf,huang2022stylizednerf,fan2022unified}.

\subsection{Gaussian Splatting}

Point-based rendering approaches have made the rendering process of point clouds differentiable, enabling end-to-end training for novel view synthesis. 3D Gaussian Splatting (3D-GS)~\cite{kerbl2023d} introduced scene modeling using 3D Gaussian distributions, compositing the appearance of intersected Gaussians along viewing rays through $\alpha$-blending.
This seminal work has inspired numerous extensions across various applications, including dynamic scene view synthesis~\cite{luiten2023dynamic,yang2024deformable,huang2024sc}, 3D content generation~\cite{zou2024triplane,tang2023dreamgaussian}, geometry reconstruction~\cite{lyu20243dgsr,huang20242d}, video representation~\cite{sun2024splatter} and neural network-enhanced high-fidelity view synthesis~\cite{yang2024spec,lu2024scaffold}. The flexible representation, computational efficiency, and superior rendering quality of 3D-GS have established it as a prominent approach in the field.

The key innovation of 3D-GS lies in its use of Gaussian kernels for point effect decay, where the continuous nature of these functions facilitates the optimization of discrete representations. Subsequent research has focused on both constraining and extending these Gaussian kernels for specific applications. Notable developments include the constraint of 3D Gaussians to 2D space~\cite{guedon2024sugar,huang20242d,dai2024high}, yielding planar representations that better approximate scene surfaces for precise geometry reconstruction. Other works have extended the framework to 4D space~\cite{duan20244d,yang2023real}, enabling temporal dynamic modeling through 3D slicing of 4D Gaussians for view synthesis. 
To address limitations in representing sharp edges, researchers have proposed modifications to the Gaussian shape using curve-cutting~\cite{qu2024disc}, hemispheric truncation~\cite{li20243d}, and novel 3D convex shapes~\cite{Held20243DConvex}.

\section{Preliminaries}
\label{subsec:method_pre}

\cyp{In this section, we first review 3D Gaussian Splatting (3D-GS) and 2D Gaussian Splatting (2D-GS), and their mathematical formulations (Sec.~\ref{subsec:pre_3dgs} and Sec.~\ref{subsec:pre_2dgs}). We then introduce our general kernel splatting framework that extends beyond traditional Gaussian primitives (Sec.~\ref{subsec:pre_gks}). Finally, we analyze the inherent limitations of Gaussian-based representations, motivating the need for our proposed deformable radial kernel formulation (Sec.~\ref{subsec:pre_discussion}).}

\subsection{3D Gaussian Splatting \cyp{(3D-GS)}}
\label{subsec:pre_3dgs}

\cyp{3D Gaussian Splatting represents scenes using colored 3D Gaussians~\citep{kerbl2023d}. Each Gaussian $G$ is parameterized by a 3D center $\mu$, covariance matrix $\Sigma = RSS^TR^T$ (where $R$ is a rotation matrix from quaternion $q \in \mathbf{SO}(3)$ and $S$ is a scaling matrix from vector $s$), opacity $o$, and spherical harmonic coefficients $sh$ for view-dependent appearance. A scene is represented as $\mathcal{G} = \{G_j: \mu_j, q_j, s_j, o_j, sh_j \}$.}

\cyp{Rendering involves projecting Gaussians onto the image plane with 2D covariance $\Sigma^{\prime} = JW\Sigma W^{T}J^{T}$ and center $\mu^{\prime} = JW\mu$. The pixel color $C(u)$ is computed via neural point-based $\alpha$-blending:}
\begin{equation}
\vspace{-2mm}
\label{equa:gaussian_render}
\small
C ({u}) = \sum_{i \in N} T_i \alpha_i \mathcal{SH}(sh_i, v_i), \text{ where } T_i = \Pi_{j=1}^{i - 1}(1 - \alpha_{j}),
\end{equation}
\cyp{where $\mathcal{SH}$ evaluates spherical harmonics with view direction $v_i$, and $\alpha_i$ is:}
\begin{equation}
\small
\label{equa:render2}
\alpha_i = o_i e^{-\frac{1}{2} ({p} - \mu_i^{\prime})^T \Sigma_i^{\prime} ({p} - \mu_i^{\prime}) }.
\end{equation}
\cyp{By optimizing these Gaussian parameters and adaptively adjusting their density, 3D-GS achieves high-quality scene representation with real-time rendering capabilities.}

\subsection{2D Gaussian Splatting \cyp{(2D-GS)}}
\label{subsec:pre_2dgs}
\cyp{Recognizing that real-world scenes primarily consist of surface structures where 3D Gaussians naturally compress into planar formations, recent works~\citep{huang20242d,dai2024high} proposed 2D Gaussian Splatting (2D-GS). The opacity $\alpha$ is computed using local coordinates $(u, v)$ on the tangent plane:}
\begin{equation}
\label{eq:2dgs}
\alpha = o \cdot \exp\left(-\frac{1}{2}\left(\frac{u^2}{s_u^2} + \frac{v^2}{s_v^2}\right)\right),
\end{equation}
\cyp{where $s_u^2$ and $s_v^2$ denote the variances along the U and V axes. This surface-centric formulation aligns with the success of polygon meshes in 3D modeling, offering a more compact scene representation.}

\begin{figure*}[t]
\begin{center}  
\includegraphics[width=.9\linewidth]{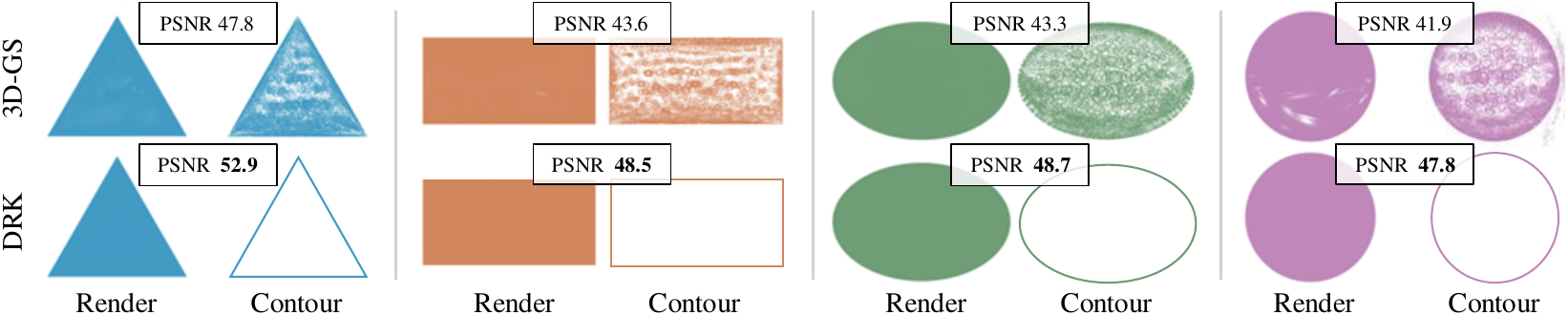}
\vspace{-5mm}
\end{center}
\caption{Comparison of 3D-GS versus a single \textit{DRK}: \textit{DRK} achieves superior geometric fidelity with just one primitive compared to multiple Gaussians. We visualize the contours of 3D-GS and \textit{DRK} to better illustrate primitive count, scale, and position.}
\label{fig:demo}
\vspace{-5mm}
\end{figure*}

\subsection{General Kernel Splatting}
\label{subsec:pre_gks}
\begin{figure}[H]
\begin{center}
\vspace{-3mm}
\includegraphics[width=\linewidth]{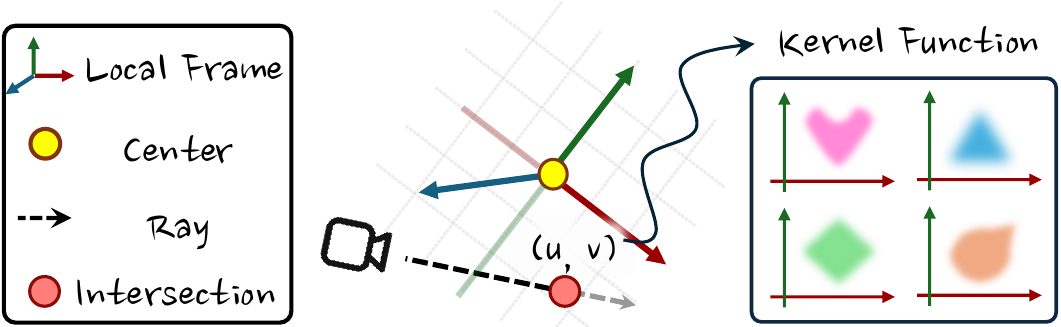}
\vspace{-10mm}
\end{center}
\caption{Illustration of general planar kernel splatting: UV coordinates of ray-plane intersections are mapped to density values via a kernel function that determines the primitive's shape.}
\label{fig:projection}
\end{figure}

\cyp{
We extend conventional 2D Gaussians by introducing a more general planar kernel splatting formulation (Fig.~\ref{fig:projection}). Given a ray from position $r_o \in \mathbb{R}^3$ with direction $r_d \in \mathbb{R}^3$ intersecting a tangent plane centered at $\mu$ with rotation $R \in SO(3)$, the intersection point is:}
\begin{equation}
\small
i = r_o + \frac{(\mu-r_o)^T R_z}{r_d^T R_z}r_d,
\end{equation}
\cyp{where $R_z$ is the plane normal. The local UV coordinates at this intersection are:}
\begin{equation}
\small
\begin{pmatrix}
u \\
v
\end{pmatrix}
=
\begin{pmatrix}
R_x^T \\
R_y^T
\end{pmatrix}
(i-\mu),
\end{equation}
\cyp{where $R_x^T$ and $R_y^T$ are the first and second rows of $R$. The kernel splatting function then maps these UV coordinates to a density value $\alpha$, determining the primitive's shape.
}

\subsection{Discussions}
\label{subsec:pre_discussion}
\cyp{
While 3D-GS demonstrates remarkable efficacy, our analysis reveals several inherent limitations. \emph{First}, screen-space projection of Gaussians yields 2D elliptical distributions with \textbf{rotational symmetry}, constraining their ability to approximate diverse primitive shapes. \emph{Second}, their \textbf{conic curve} boundaries (derived by the $L2$ norm) challenge the representation of linear edges and intermediate edge forms. \emph{Third}, Gaussian distributions inherently couple \textbf{decay rate} with spatial extent through the covariance matrix—sharp features necessitate narrow distributions, making it challenging to simultaneously capture abrupt transitions and extended spatial regions.}

\begin{figure}[hb]
\begin{center}
\vspace{-3mm}
\includegraphics[width=\linewidth]{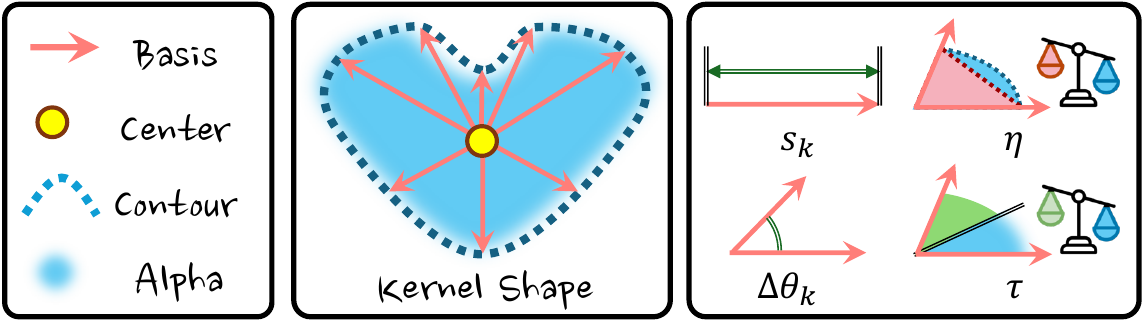}
\vspace{-10mm}
\end{center}
\caption{\textit{DRK} defines a deformable shape using radial basis functions characterized by lengths $s_k$ and polar angles $\theta_k$, with parameters $\eta$ and $\tau$ governing the shape's curvature and sharpness respectively.}
\label{fig:kernel}
\vspace{-5mm}
\end{figure}

\cyp{These constraints necessitate millions of fine-grained Gaussians to approximate arbitrary shapes, leading to over-parameterization while still failing to achieve perfect fidelity under practical constraints. As shown in Fig.~\ref{fig:demo}, even elementary shapes like triangles and rectangles—which can be described concisely—require numerous Gaussians for approximation, exhibiting artifacts like interior inconsistencies and external floating points (last column). Our deformable radial kernel (DRK) addresses these limitations through radial basis functions with varying lengths $s_k$ and polar angles $\theta_k$, controlled by parameters $\eta$ and $\tau$ for contour curvature and sharpness (Fig.~\ref{fig:kernel}).
}

\section{Deformable Radial Kernel Splatting}

\cyp{Deformable Radial Kernel (DRK) is a novel primitive for 3D scene representation characterized by a set of parameters $\Theta = \{\mu, q, s_k, \theta_k, \eta, \tau, o, sh\}$, where:
\begin{itemize}
    \item $\mu \in \mathbb{R}^3$, $q \in SO(3)$, $o$, and $sh$ follow 3D-GS for position, orientation, opacity, and view-dependent appearance
    \item $\{s_k, \theta_k\}_{k=1}^K$ define radial basis lengths and angles that control the kernel shape
    \item $\eta \in (0,1)$ and $\tau \in (-1,1)$ control boundary curvature and sharpness, respectively
\end{itemize} }

In the following sections, we detail each component of DRK. First, we introduce the radial basis formulation (Sec.~\ref{subsec:radial}), showing that traditional 2D Gaussians are a special case within our framework. Then, we discuss $L1\&L2$ norm blending and edge sharpening in Sec.~\ref{subsec:l1l2} and Sec.~\ref{subsec:sharpen}, which enable flexible control of boundary curvature and sharpness. Finally, in Sec.~\ref{subsec:rasterization}, we explore efficient rasterization strategies specifically tailored for DRK, including low-pass filtering, kernel culling, and cache sorting.

\subsection{Radial Basis}
\label{subsec:radial}
\cyp{
Our radial basis defines a kernel's shape through $K$ control points in polar coordinates $\{ s_k, \theta_k \}$, where $s_k > 0$ represents each radial length and angles are ordered as $0 < \theta_0 < ... < \theta_K \leq 2\pi$. For any point with local tangent coordinates $(u,v)$, we compute its polar representation using $L2$ norm $r_2 = \sqrt{u^2 + v^2}$ and angle $\theta = \arccos(u/r_2)$. Given a point lying between basis $k$ and $k+1$ (i.e., $\theta_k < \theta \leq \theta_{k+1}$), we define its relative angular distance as $\Delta\theta_k = \frac{(\theta - \theta_k)\pi}{\theta_{k+1} - \theta_k}$. The kernel function is then formulated as:
\begin{equation}
\small
\label{eq:radial}
    \alpha = o \cdot \exp\left(- \frac{r_2^2}{2} \left( \frac{1+\cos(\Delta\theta_k)}{2s_k^2} + \frac{1-\cos(\Delta\theta_k)}{2s_{k+1}^2} \right) \right).
\end{equation}
This formulation smoothly interpolates between adjacent radial bases using cosine weighting terms, ensuring continuous transitions in the kernel's shape.
}
\vspace{-4mm}
\paragraph{2D Gaussian as a Special Case.}
\cyp{
Our formulation generalizes 2D Gaussian kernels. By setting $K=4$ with $\theta_k = \frac{k\pi}{2}$ and $s_0=s_2=s_u$, $s_1=s_3=s_v$, we can derive the 2D Gaussian form in Eq.~\eqref{eq:2dgs}. For points in the first UV quadrant where $\Delta\theta_k=2\theta$:
\begin{align*} 
\small
\alpha
&= o \cdot \exp\left(- \frac{r_2^2}{2} \left( \frac{\cos^2\theta}{s_0^2} + \frac{\cos^2\theta}{s_1^2} \right) \right) \\
&= o \cdot \exp\left(- \frac{1}{2} \left( \frac{u^2}{s_0^2} + \frac{v^2}{s_1^2} \right) \right),
\end{align*}
where we use the identity $\cos(\Delta\theta_k)=\cos(2\theta)=2\cos^2\theta-1$. The derivation extends to other quadrants by symmetry.
}

\subsection{\cyp{$L1\&L2$} Norm Blending}
\label{subsec:l1l2}

\cyp{The scaling term $\frac{1}{\bar{s}^2} = \left( \frac{1+\cos(\Delta\theta_k)}{2s_k^2} + \frac{1-\cos(\Delta\theta_k)}{2s_{k+1}^2} \right)$ in Eq.~\eqref{eq:radial} enforces smooth contours at radial endpoints, where $\frac{d\bar{s}}{d\Delta\theta_k}=0$ at $\Delta\theta_k=0$ or $\pi$. However, since radials converge at the kernel center and are not parallel, this $L2$ formulation inherently produces conic curves, limiting its ability to represent \textbf{straight edges} common in man-made environments.}

\cyp{To address this limitation, we incorporate $L1$ norm into our \textit{DRK} formulation. For a point $(u,v)$ between radial endpoints $e_i = (s_i\cos(\theta_i), s_i\sin(\theta_i))$ and $e_{i+1}$, we compute its $L1$ norm as:
\begin{equation}
\small
\label{eq:L1}
    r_1 = \left\|
    \begin{pmatrix}
        e_i & e_{i+1}
    \end{pmatrix}^{-1}
    \begin{pmatrix}
        u \\ v
    \end{pmatrix}\right\|_1.
\end{equation}
This transformation maps the diamond-shaped $L1$ unit ball ($|x|+|y|=1, 0\leq x,y \leq 1$) to the segment between adjacent endpoints, enabling straight-line boundaries.}

\cyp{We then introduce a blending weight $\eta \in (0,1)$ to smoothly interpolate between $L1$ and $L2$ norms, yielding our complete kernel function:
\begin{equation}
\small
\label{eq:kernel}
    \alpha = o \cdot \exp\left( -\frac{1}{2} \left( \eta r_1^2 + (1-\eta)\frac{r_2^2}{\bar{s}^2} \right) \right).
\end{equation}
Note that $r_1^2$ requires no explicit scaling term as it is inherently scaled through the inverse transformation in Eq.~\eqref{eq:L1}. Please see Fig.~\ref{fig:kernel} (right) for illustration.} 

\subsection{Edge Sharpening}
\label{subsec:sharpen}

\begin{wrapfigure}{r}{0.17\textwidth}
    \centering
    \vspace{-4mm}
    \hspace{-10mm}
    \hspace{3mm}
    \includegraphics[width=0.2\textwidth]{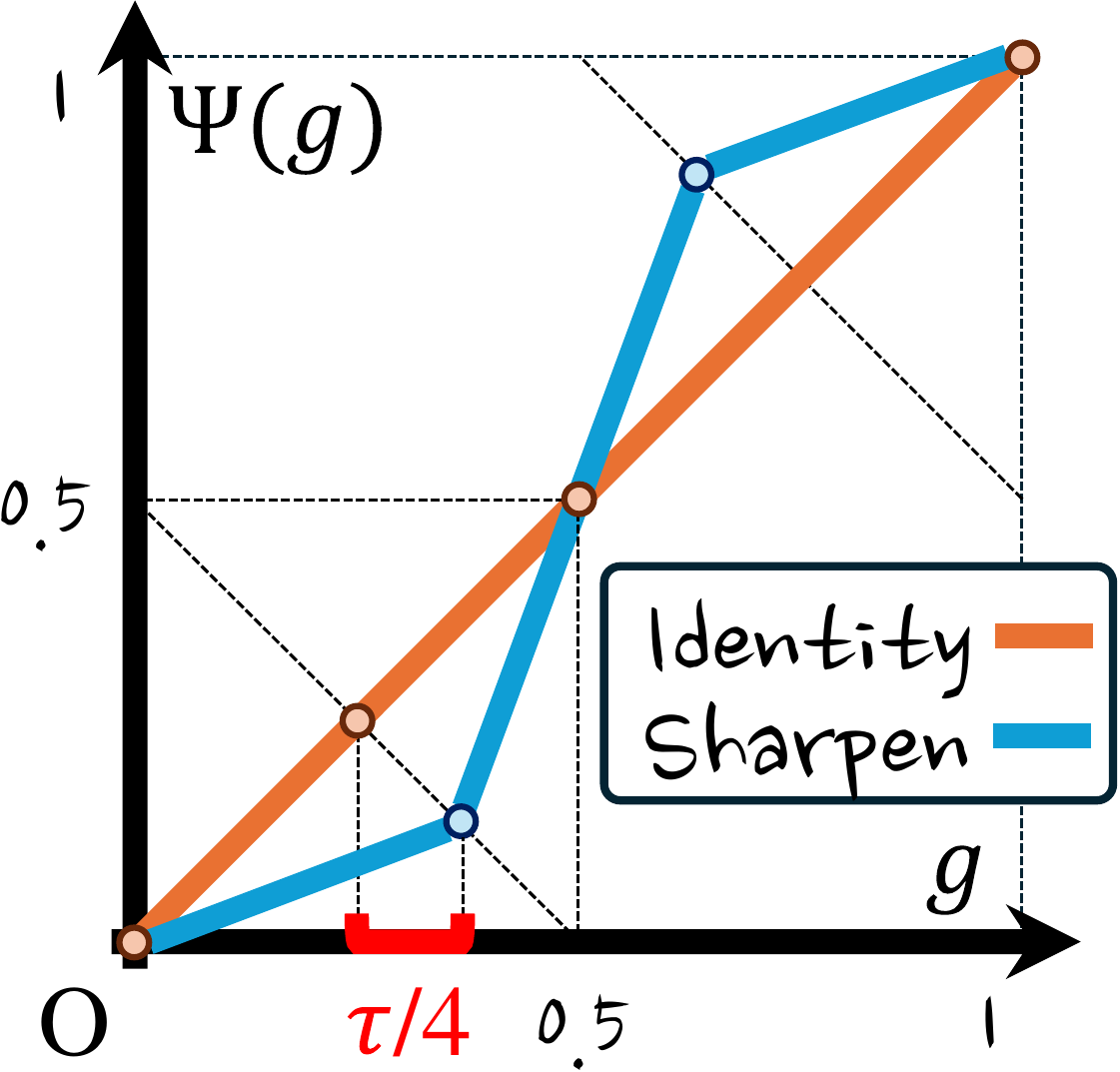}
    \vspace{-6mm}
    \caption{Sharpening function illustration.}
    \label{fig:sharpen}
    \vspace{-3mm}
\end{wrapfigure}

\cyp{
As shown in Fig.~\ref{fig:demo}, Gaussian functions inherently couple scale and edge sharpness through their variance parameter, making it challenging to achieve precise shape approximation even with thousands of Gaussians. To decouple these properties, we introduce a sharpening coefficient $\tau \in (-1, 1)$ that adaptively adjusts edge transitions. Given the exponential term $g$ (i.e., the density value before opacity scaling) in Eq.~\eqref{eq:kernel}, our sharpening function is:
\begin{equation}
\small
    \Psi(g) = 
    \left\{
    \begin{array}{llcl}
                \frac{1-\tau}{1+\tau} g                        & \text{if } &  0                & \leq g < \frac{1+\tau}{4},\\
                \frac{1+\tau}{1-\tau} g - \frac{\tau}{1-\tau}  & \text{if } &  \frac{1+\tau}{4} & \leq g < \frac{3-\tau}{4},\\
                \frac{1-\tau}{1+\tau} g + \frac{2\tau}{1+\tau} & \text{if } &  \frac{3-\tau}{4} & \leq g \leq 1.
    \end{array}
    \right.
\end{equation}
As illustrated in Fig.~\ref{fig:sharpen}, this piecewise function reshapes the alpha values toward 0 or 1, creating sharper edge transitions while maintaining the kernel's spatial extent. The final kernel opacity is computed as $\alpha = o \cdot \Psi(g)$.
}

\subsection{Rasterization}
\label{subsec:rasterization}
\paragraph{Low-pass Filtering.}
\cyp{Camera-captured images represent discretely sampled signals from scene rays, inherently limiting the maximum recoverable frequency of 3D content. Without proper frequency constraints, optimization tends to generate small, floating primitives that overfit to individual training views. These primitives become too fine-grained (smaller than ray coverage) and only contribute to specific rays, resulting in poor generalization.}

\cyp{While 3D-GS leverages EWA splatting for closed-form Gaussian low-pass filtering, \textit{DRK} requires an alternative approach. Following Botsch et al.~\cite{botsch2005high}, we approximate filtering by taking the maximum between the kernel function and a view-dependent low-pass filter. Since \textit{DRK} represents planar primitives, its frequency should be bounded on its tangent plane while potentially unbounded when viewed orthogonally, so we scale the filter size by the cosine of the view angle:}
\begin{equation}
\small
\Tilde{\alpha} = \max(\alpha, o \cdot \text{exp}(- \frac{1}{2 |r_d\cdot R_z|^2} (\frac{\Delta p_w^2}{s_l^2} +\frac{\Delta p_h^2}{s_l^2}) )),
\end{equation}
where $\Delta p_h$ and $\Delta p_w$ are image space distances between the pixel and the projected kernel center, $r_d\cdot R_z$ accounts for view-dependent scaling, and $s_l$ controls the filter radius.

\begin{table*}[t]
  \centering
  \setlength\tabcolsep{5pt}
  \resizebox{1\linewidth}{!}{
    \begin{tabular}{@{\extracolsep{\fill}} l|SSS|SSS|SSS|SSS|SSS|SSSS }
        \toprule[1pt]
        \multirow{2}{*}{Methods} &
          \multicolumn{3}{c}{\textbf{Simple}} &
          \multicolumn{3}{c}{\textbf{Texture}} &
          \multicolumn{3}{c}{\textbf{Geometry}} &
          \multicolumn{3}{c}{\textbf{Specular}} &
          \multicolumn{3}{c}{\textbf{Large}} &
          \multicolumn{3}{c}{\textbf{Average}} \\
          \cmidrule(lr){2-4}
          \cmidrule(lr){5-7}
          \cmidrule(lr){8-10}
          \cmidrule(lr){11-13}
          \cmidrule(lr){14-16}
          \cmidrule(lr){17-19}
          & {\textit{\footnotesize{PSNR}}} & {\footnotesize{\textit{LPIPS}}} & {\footnotesize{\textit{SSIM}}} 
          & {\footnotesize{\textit{PSNR}}} & {\footnotesize{\textit{LPIPS}}} & {\footnotesize{\textit{SSIM}}}
          & {\footnotesize{\textit{PSNR}}} & {\footnotesize{\textit{LPIPS}}} & {\footnotesize{\textit{SSIM}}}
          & {\footnotesize{\textit{PSNR}}} & {\footnotesize{\textit{LPIPS}}} & {\footnotesize{\textit{SSIM}}}
          & {\footnotesize{\textit{PSNR}}} & {\footnotesize{\textit{LPIPS}}} & {\footnotesize{\textit{SSIM}}}
          & {\footnotesize{\textit{PSNR}}} & {\footnotesize{\textit{LPIPS}}} & {\footnotesize{\textit{SSIM}}} \\
          \midrule
            2D-GS
            & \cellcolor{third}{42.10} & {.0209} & \cellcolor{third}{.9969}
            & {42.59} & {.0624} & {.9834}
            & {29.90} & {.0513} & {.9700}
            & \cellcolor{third}{27.18} & \cellcolor{second}{.0570} & {.9550}
            & {27.80} & {.2500} & {.8388}
            & {33.92} & {.0881} & {.9514} \\
            3D-GS
            & {39.51} & {.0311} & {.9941}
            & {44.57} & {.0628} & {.9968}
            & {29.14} & {.0552} & {.9658}
            & {27.16} & {.0758} & {.9523}
            & {32.19} & {.2054} & \cellcolor{third}{.9017}
            & {34.41} & {.0861} & {.9621} \\
            3D-HGS  
            & {41.41} & \cellcolor{third}{.0199} & {.9931}
            & {46.19} & \cellcolor{second}{.0397} & {.9912}
            & {30.42} & \cellcolor{third}{.0406} & {.9598}
            & {27.27} & \cellcolor{best}{.0512} & {.9447}
            & \cellcolor{third}{33.17} & \cellcolor{second}{.1672} & {.8721}
            & \cellcolor{third}{35.68} & \cellcolor{second}{.0637} & {.9521} \\
            GES
            & {40.95} & {.0266} & {.9955}
            & {45.23} & {.0579} & \cellcolor{third}{.9977}
            & {30.36} & {.0468} & {.9704}
            & {27.04} & {.0632} & {.9550}
            & {31.67} & {.2074} & {.8987}
            & {35.05} & {.0804} & {.9634} \\
            \hline
            \textit{DRK} (S2)
            & {40.34} & {.0283} & {.9946}
            & \cellcolor{third}{47.67} & {.0537} & {.9976}
            & \cellcolor{third}{30.68} & {.0477} & \cellcolor{third}{.9709}
            & {27.07} & {.0704} & \cellcolor{third}{.9570}
            & {31.57} & {.2090} & {.8985}
            & {35.03} & {.0823} & \cellcolor{third}{.9637} \\
            \textit{DRK} (S1)
            & \cellcolor{second}{42.24} & \cellcolor{second}{.0168} & \cellcolor{second}{.9974}
            & \cellcolor{second}{47.86} & \cellcolor{third}{.0410} & \cellcolor{second}{.9988}
            & \cellcolor{second}{31.67} & \cellcolor{second}{.0394} & \cellcolor{second}{.9788}
            & \cellcolor{second}{27.77} & {.0631} & \cellcolor{second}{.9603}
            & \cellcolor{second}{33.78} & \cellcolor{third}{.1685} & \cellcolor{second}{.9196}
            & \cellcolor{second}{36.62} & \cellcolor{third}{.0668} & \cellcolor{second}{.9701} \\
            \textit{DRK}
            & \cellcolor{best}{43.46} & \cellcolor{best}{.0105} & \cellcolor{best}{.9985}
            & \cellcolor{best}{49.08} & \cellcolor{best}{.0396} & \cellcolor{best}{.9990}
            & \cellcolor{best}{31.93} & \cellcolor{best}{.0366} & \cellcolor{best}{.9803}
            & \cellcolor{best}{28.19} & \cellcolor{third}{.0592} & \cellcolor{best}{.9615}
            & \cellcolor{best}{35.28} &\cellcolor{best}{.1348} & \cellcolor{best}{.9372}
            & \cellcolor{best}{37.58} & \cellcolor{best}{.0564} & \cellcolor{best}{.9752} \\
            
        \bottomrule[1pt]
      \end{tabular}
      }
  \vspace{-3mm}
  \caption{Rendering quality evaluation on DiverseScene datasets across various categories}\label{tab:synthetic}
    \vspace{-4mm}
\end{table*}

\begin{table}[ht]
  \centering
  \adjustbox{width={\linewidth},keepaspectratio}{
    \begin{tabular}{lcccc|ccc}
        \toprule[1pt]
        & {2DGS} & {3DGS} & {3DHGS} & {GES} & {\textit{DRK}(S2)} & {\textit{DRK}(S1)} & {\textit{DRK}} \\
        \cmidrule(r){1-8} %
        \textbf{Num($\downarrow$)} 
        & {359K} & {336K} & {373K} & {330K} & \cellcolor{best}{42 K} & \cellcolor{second}{109K} & \cellcolor{third}{260K} \\
        \textbf{MB($\downarrow$)} 
        & {83.6} & {79.7} & {89.6} & {78.1} & \cellcolor{best}{12.3} & \cellcolor{second}{32.1} & \cellcolor{third}{76.6} \\
        \textbf{FPS($\uparrow$)} 
        & \cellcolor{best}{251.3} & \cellcolor{second}{247.1} & {154.5} & {227.4} & \cellcolor{third}{234.9} & {119.2} & {77.5} \\
        \bottomrule[1pt]
    \end{tabular}}
  \vspace{-3mm}
  \caption{Evaluation results of average primitive number (Num), model size (MB), and rendering speed (FPS) on DiverseScene.}\label{tab:syn_extra}
    \vspace{-0mm}
\end{table}

\begin{figure}[thb]
\begin{center}
\vspace{-3mm}
\includegraphics[width=\linewidth]{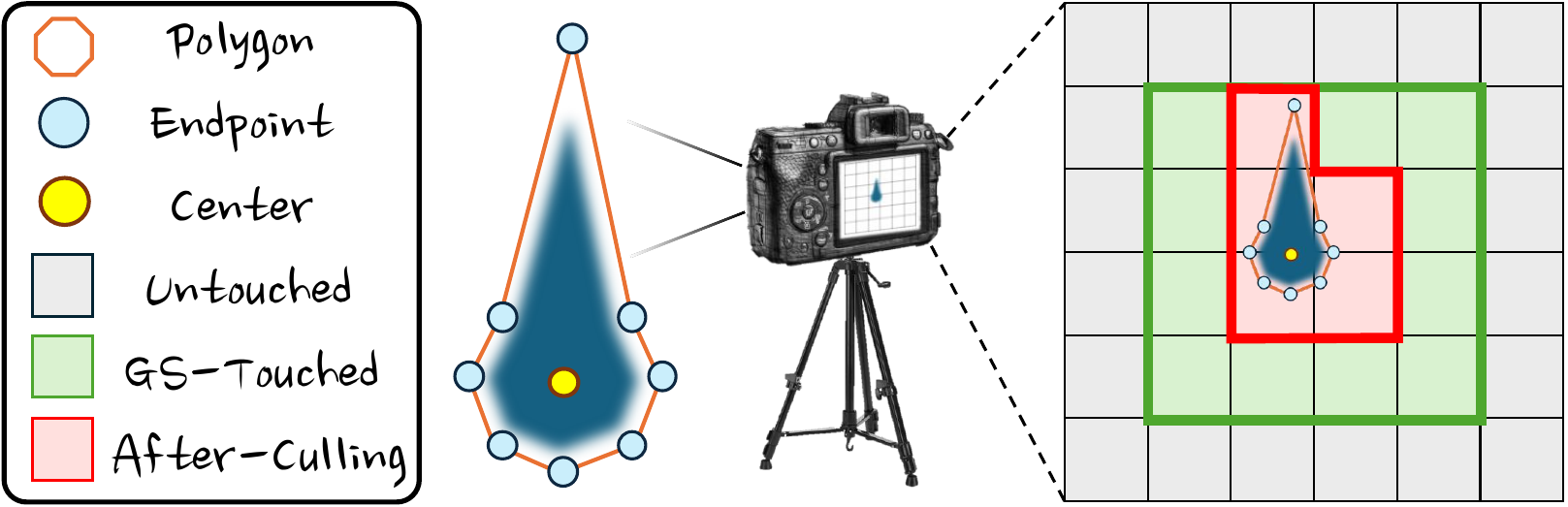}
\vspace{-10mm}
\end{center}
\caption{Polygon-based tile culling using radial basis endpoints enables efficient rasterization by eliminating untouched tiles.}
\label{fig:culling}
\vspace{-3mm}
\end{figure}

\paragraph{Kernel Culling.}
Our rendering pipeline, based on 3D-GS, performs image rendering using 2D thread grids on CUDA. The process begins by dividing images into 16×16 tiles, with each tile corresponding to potentially intersecting primitives (kernels or 3D Gaussians). 
While 3D-GS projects the major axis of a 3D Gaussian onto the image plane to create a square axis-aligned bounding box (AABB) for tile coverage, this method can be computationally inefficient for primitives that project as highly anisotropic or elongated shapes, as shown in Fig.~\ref{fig:culling}. This inefficiency becomes particularly evident in \textit{DRK} 
\cyp{due to its diverse shape representations.}
To address this, we introduce a more precise radius calculation and an improved tile culling strategy.

The effective range of \textit{DRK} is governed by its radial basis, whose endpoints define the kernel's boundary. Following the 3-$\sigma$ principle, we define the boundary radius $s^c_k$, but extend it to account for opacity $o$ and sharpness $\tau$ parameters. The calibrated radial length $s_k^c$ is defined as:
\begin{equation}
s^c_k = s_k \sqrt{ - \log \left( \Psi^{-1}(\frac{e^{-3^2}}{o}) \right)}.
\label{eq:calibrate}
\end{equation}
The derivation details are provided in the Supplementary Material. Using this calibrated length, we calculate each radial basis endpoint $v_k$:
\begin{equation}
v_k = \mu + s_k^c (\cos\theta_k R_x + \sin\theta_k R_y).
\end{equation}
These endpoints form a polygon that accurately approximates the kernel's boundary. By projecting this polygon onto the image plane, we can efficiently identify and cull tiles that don't intersect with the kernel, making the rasterization more efficient.

\begin{figure}[thb]
\begin{center}
\vspace{-3mm}
\includegraphics[width=\linewidth]{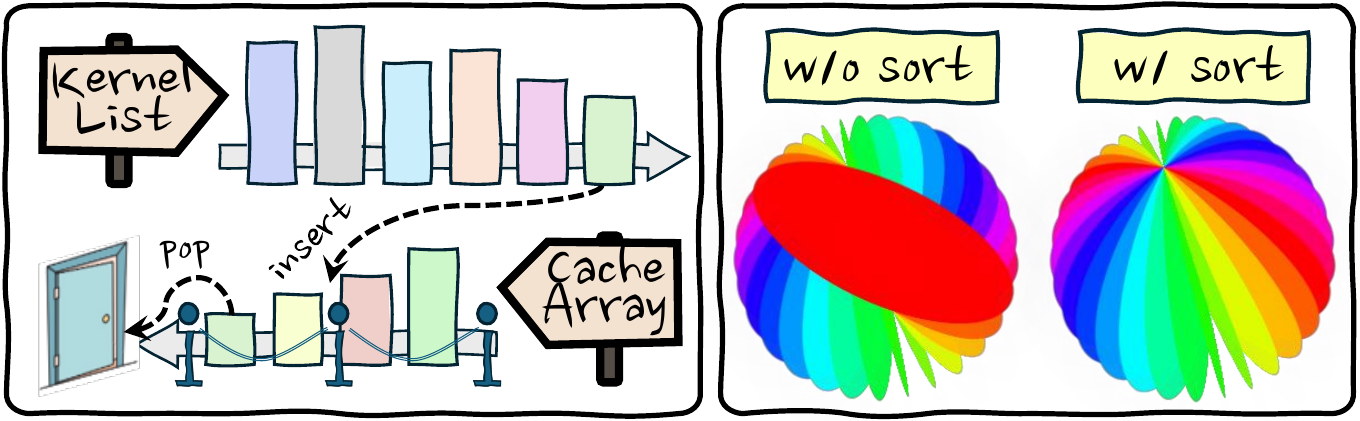}
\vspace{-11mm}
\end{center}
\caption{Our cache-sorting uses $r_t$ as keys in cache-sorting to rectify the sorting order while preventing popping artifacts.}
\label{fig:cache-sort}
\vspace{-6mm}
\end{figure}

\paragraph{Cache Sorting.}
3D-GS sorts the Gaussian using the depth of the center $\mu$, which can lead to inconsistencies across multiple views. \cyp{While} Radl et al.~\cite{radl2024stopthepop} address this by estimating the depth of the Gaussian at the point of maximum density and applying a hierarchical sorting solution to mitigate popping artifacts caused by inconsistent sorting, \cyp{we propose a simpler yet effective solution.}
In our approach, the distance $r_t$ from the camera to the intersection between the ray and \textit{DRK} is calculated as:
\begin{equation}
r_t = \frac{(\mu-r_o)^T R_z}{r_d^T R_z}.
\end{equation}

During rendering, each thread processes a pixel using an \cyp{sorted} array to cache the $r_t$ values and kernel indices, as illustrated in Fig.~\ref{fig:cache-sort}. The array is kept sorted by performing insert sorting for each new-coming kernel. If the array is full, the kernel with the smallest $r_t$ is popped out for processing. A dynamic cursor is used to record the scanning. We maintain an array length of 8, which is sufficient for approximating a more accurate sort. Fig.~\ref{fig:cache-sort} demonstrates our method's effectiveness in handling multiple overlapping kernels with identical center depths, comparing rasterization results with and without cache-sorting, \cyp{preventing popping artifacts while maintaining rendering efficiency.}

\begin{figure*}[ht]
    \centering
    \includegraphics[width=0.95\linewidth]{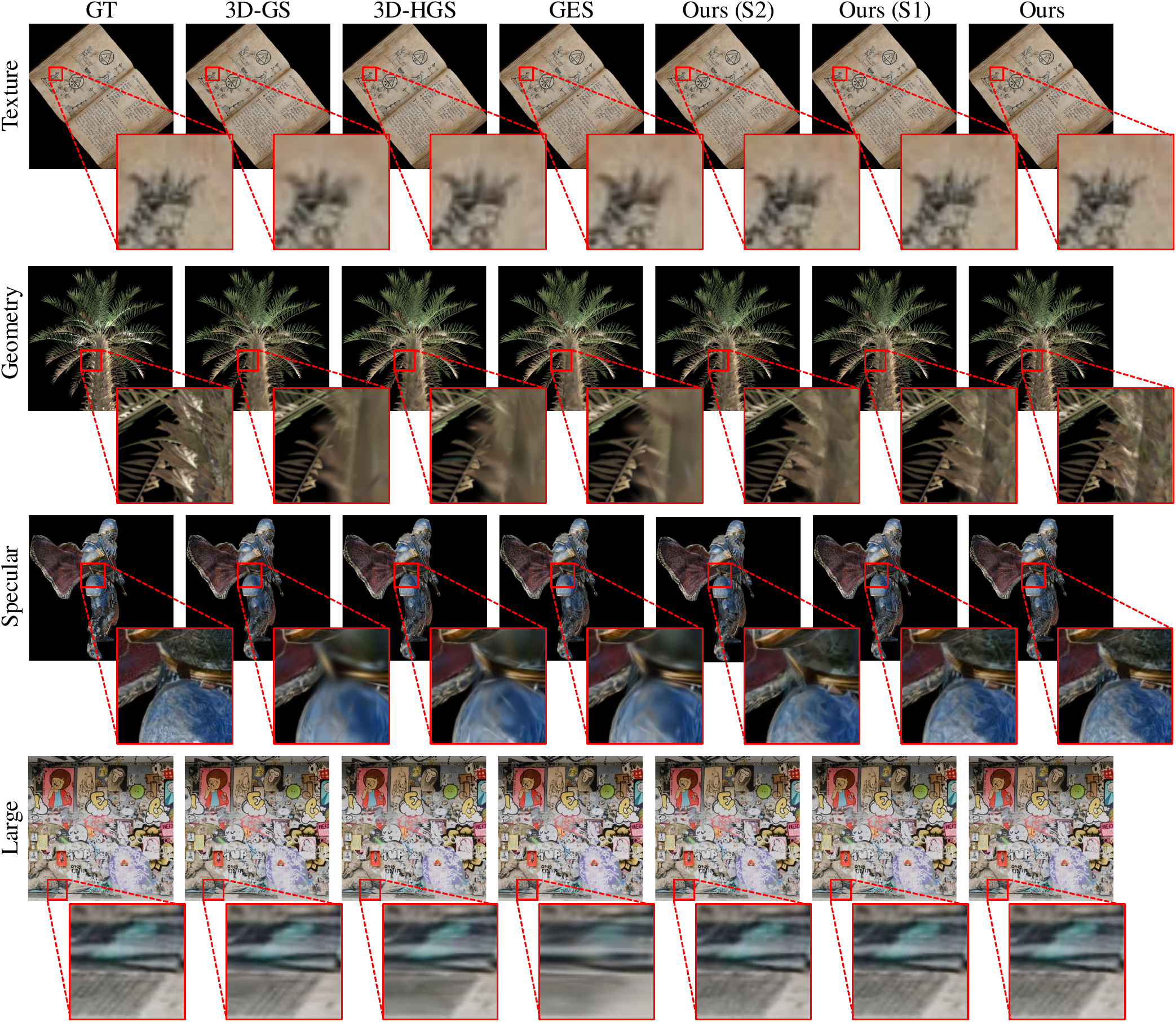}
    \vspace{-3mm}
    \caption{Qualitative comparisons of \textit{DRK} with state-of-the-art methods across various scenarios show that \textit{DRK} effectively captures sharp texture and geometry boundaries.}
    \vspace{-6mm}
    \label{fig:comparison_synthetic}
\end{figure*}

\section{Experiment}
\label{sec:experiment}

\begin{figure*}[ht]
    \centering
    \vspace{-6mm}
    \includegraphics[width=0.98\linewidth]{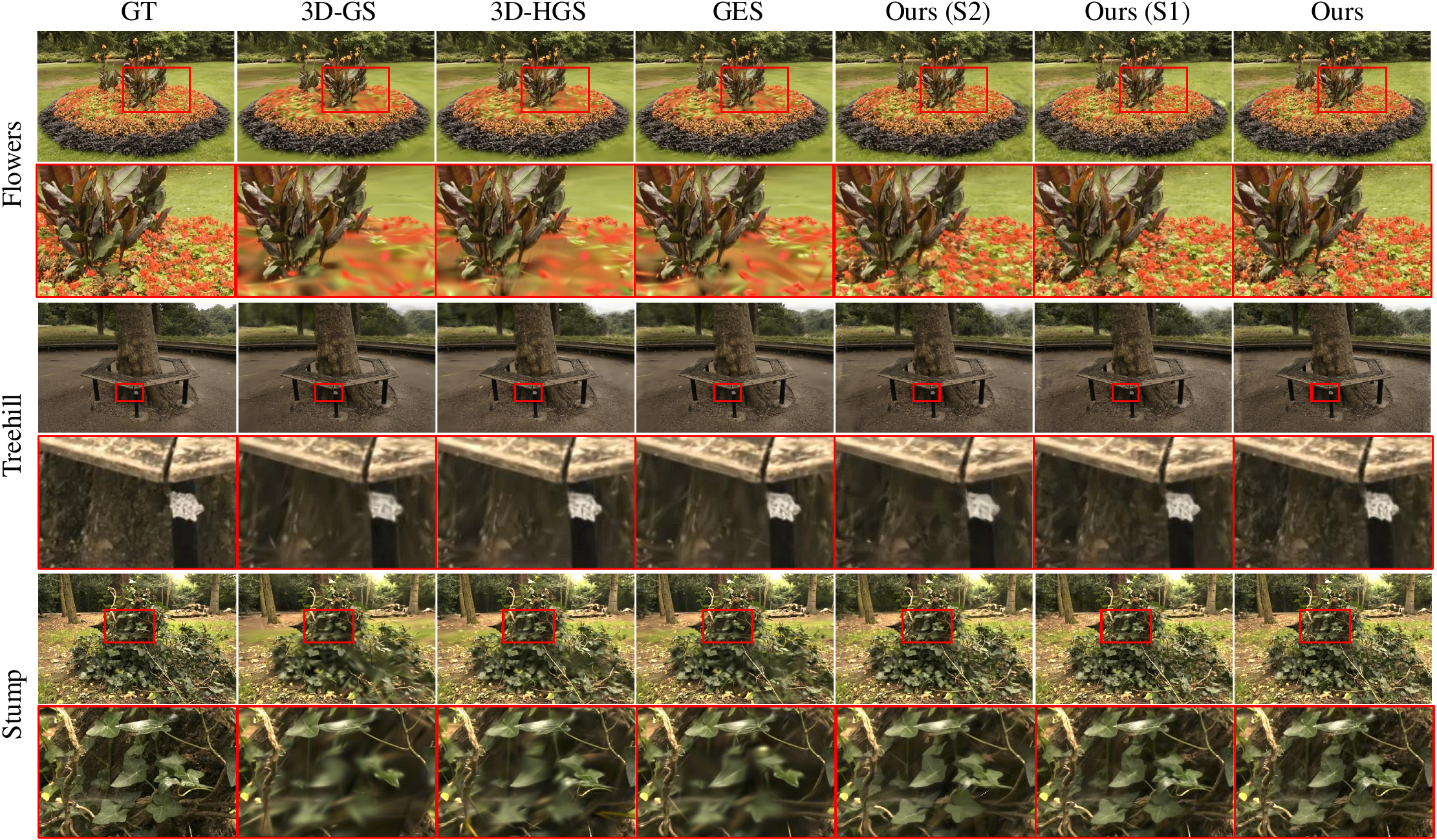}
    \vspace{-2mm}
    \caption{Comparisons on the MipNeRF 360 dataset show our method achieves clearer details and high-fidelity rendering.}
    \label{fig:comparison_mip}
    \vspace{-4mm}
\end{figure*}

\subsection{Datasets and Evaluation Metrics}
To evaluate our method's performance across diverse scenarios, we focused on four key aspects: texture complexity, geometric detail, view-dependent effects, and scene scale. We collected 10 representative 3D scenes from Sketchfab\footnote{https://sketchfab.com/}, with two scenes per category: simple objects, richly textured surfaces, intricate geometric details, specular materials, and large-scale environments. The dataset, namely \emph{DiverseScenes}, serves as a benchmark to assess our representation method's capabilities. Detailed information about DiverseScenes can be found in the supplementary materials.
For the evaluation on unbounded real-world scenes, we utilized the Mip-NeRF360 dataset~\cite{barron2022mip}, which contains multi-view images captured in open environments. Camera poses were estimated using COLMAP~\cite{schoenberger2016sfm}. We assessed performance using three metrics: Peak Signal-to-Noise Ratio (PSNR) for overall reconstruction accuracy, Structural Similarity Index (SSIM) for perceptual quality and Learned Perceptual Image Patch Similarity (LPIPS)~\cite{zhang2018perceptual} for human-perceived visual fidelity.
We observed that the number of primitives significantly impacts rendering quality. To evaluate DRK's performance across different densities, we present results using various hyper-parameters (details provided in the Supp), where the density decreases progressively from \textit{DRK} to \textit{DRK}(S2) and then to \textit{DRK}(S1).

\subsection{Quantitative Comparisons}

\noindent\textbf{DiverseScenes.} We compare our method, \textit{DRK}, against state-of-the-art Gaussian representations: 3D-GS \cite{kerbl2023d}, 2D-GS \cite{huang20242d}, 3D-HGS \cite{li20243d}, and GES \cite{hamdi2024ges}, using their official implementations. To ensure a fair comparison focused on kernel representation, we adjust the gradient threshold for densification in these methods, aligning their primitive count with that of 3D-GS (300K+). The results for each dataset category are shown in Table \ref{tab:synthetic}, demonstrating that our method achieves the best average performance. This is followed by our sparser version, \textit{DRK} (S1). \textit{DRK} (S2) also performs comparably to other state-of-the-art methods, with significantly fewer primitives and lower space occupancy, while maintaining comparable FPS, as detailed in Table \ref{tab:syn_extra}. Although our method requires more computation, the rendering framerates remain efficient (FPS $>$ 50).

\noindent\textbf{Mip-NeRF360 Datasets.} We also tested \textit{DRK} on Mip-NeRF360 datasets, known for accurate camera poses and unbounded scenes. Results are shown in Tab.~\ref{tab:mip360}. 
On unbounded scenes, \textit{DRK} still excels in perceptual quality (LPIPS, SSIM), it shows less advantage in PSNR, especially with sparser kernels. We hypothesize that DRK may tend to overfit distant regions with limited supervision. To address this, we used MiVOS\cite{cheng2021mivos} to mask image sequences, training on original images but evaluating only the foreground. The results (M-PSNR) indicate our method's strength in well-supervised central regions.

\begin{table}[t]
  \centering
    \adjustbox{width={\linewidth},keepaspectratio}{
    \begin{tabular}{lSSSSSS }
        \toprule[1pt]
        Methods
          & {\textbf{PSNR($\uparrow$)}} & {\textbf{M-PSNR($\uparrow$)}} & {\textbf{LPIPS($\downarrow$)}} & {\textbf{SSIM($\uparrow$)}}  & {\textbf{Num($\downarrow$)}}  & {\textbf{Size($\downarrow$)}}  \\
          \cmidrule(r){1-7} %
            2D-GS
            & {26.32} & {32.01} & {.2987} & {.7617} & {928 K} & {216.0 M} \\
            3D-GS
            &  {26.48} & {32.23} & {.3045} & {.7537} & \cellcolor{third}{811 K} & \cellcolor{third}{191.3 M} \\
            3D-GS (L)
            &  \cellcolor{best}{26.94} & \cellcolor{second}{32.75} & \cellcolor{third}{.2688} & \cellcolor{second}{.7799} & {1191 K} & {281.6 M} \\
            3D-HGS
            & \cellcolor{third}{26.74} & {32.51} & {.2999} & {.7561} & {842 K} & {202.4 M} \\
            GES
            & {26.62} & {32.11} & {.3047} & {.7542} & {824 K} & {195.0 M} \\
            \cmidrule(r){1-7}  %
            \textit{DRK} (S2)
            & {26.20} & {32.31} & {.2781} & {.7601} & \cellcolor{best}{388 K} & \cellcolor{best}{125.9 M} \\
            \textit{DRK} (S1)
            & {26.40} & \cellcolor{third}{32.56} &  \cellcolor{second}{.2601} & \cellcolor{third}{.7722} & \cellcolor{second}{551 K} & \cellcolor{second}{161.9 M} \\
            \textit{DRK}
            & \cellcolor{second}{26.76} & \cellcolor{best}{32.81} & \cellcolor{best}{.2364} & \cellcolor{best}{.7871} & {952 K} & {279.1 M}  \\
        \bottomrule[1pt]
      \end{tabular}
      }
\vspace{-1mm}
  \caption{Quantitative evaluation on Mip-NeRF360 scenes~\cite{barron2022mip}.}\label{tab:mip360}
  \vspace{-4mm}
\end{table}

\subsection{Qualitative Comparison}
We conducted qualitative comparisons to highlight the advantages of DRK over other state-of-the-art kernel representations. Fig.\ref{fig:comparison_synthetic} shows results across different scene categories, where our method achieves better texture and geometry fitting with a comparable or smaller number of kernels. Fig.\ref{fig:comparison_mip} demonstrates that DRK effectively handles fine geometries, such as flowers and leaves, and detailed textures, like tags and bark, outperforming other representations that may have used more primitives (see Tab.~\ref{tab:mip360}).

\subsection{Converting Mesh to DRK}
\label{sec:mesh2drk}

\begin{figure}[thb]
\begin{center}
\vspace{-2mm}
\includegraphics[width=\linewidth]{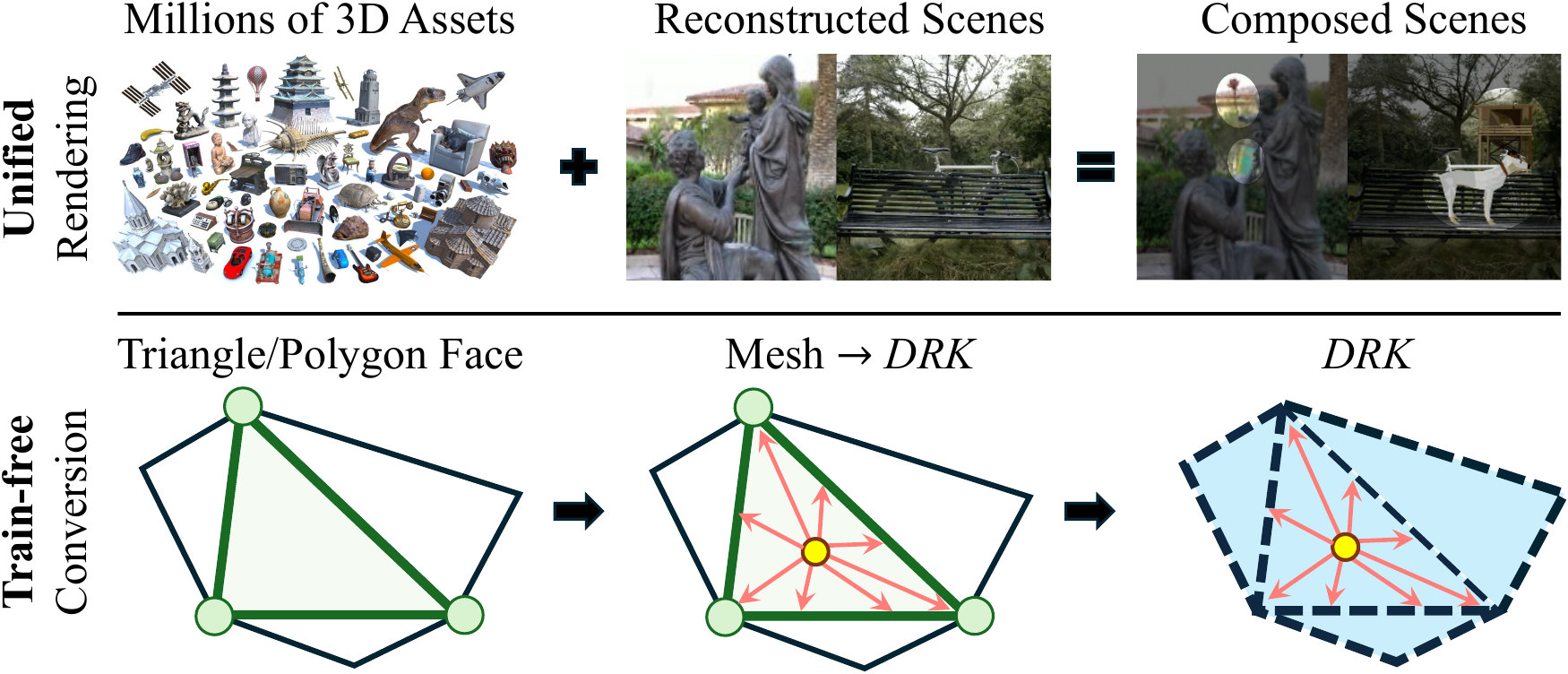}
\vspace{-10mm}
\end{center}
\caption{The conversion from Mesh to \textit{DRK} demonstrates the efficiency of using \textit{DRK} to incorporate traditional 3D assets into any \textit{DRK} scene at low cost.}
\label{fig:mesh2drk_main}
\vspace{-5mm}
\end{figure}

All existing "Splatting" methods~\cite{kerbl2023d,huang20242d,hamdi2024ges,li20243d} fail to seamlessly integrate with traditional assets. In contrast, our \textit{DRK} framework provides a versatile representation that allows for effortless conversion of triangle and polygon faces into \textit{DRK}. This is accomplished by placing the \textit{DRK} endpoints at the polygon vertices and adjusting the parameters $\tau$ and $\eta$ to control sharpness, with both $L1$ and $L2$ curvature set to 1. This process eliminates the need for training data rendering or model optimization. Consequently, rich 3D assets can be seamlessly transferred into \textit{DRK}, bridging the gap between \textbf{millions of traditional 3D assets} and our fast, high-quality reconstruction representation. Figure~\ref{fig:mesh2drk_main} illustrates the potential applications of Mesh2\textit{DRK}.

\section{Conclusion}

We present \textit{DRK}, a novel primitive representation that generalizes and enhances Gaussian splatting. By incorporating learnable radial bases, adaptive shape control, and efficient rendering strategies, \textit{DRK} effectively addresses the fundamental limitations of Gaussian kernels in representing diverse geometric features. Our experiments on both synthetic and real-world scenes demonstrate that \textit{DRK} achieves superior rendering quality with significantly fewer primitives, while maintaining computational efficiency through carefully designed rasterization optimizations. 
Leveraging the flexibility of \textit{DRK}, traditional triangle or polygon mesh representations can be seamlessly converted into \textit{DRK} without requiring any training. This capability bridges rich 3D assets with reconstructed scenes, paving the way for future applications such as real-scene editing, VR digital world creation, and AR scene enhancement.

\section*{Acknowledgement}
\noindent This work has been supported in part by Hong Kong Research Grant Council - Early Career Scheme (Grant No. 27209621), General Research Fund Scheme (Grant No. 17202422, 17212923), Theme-based Research (Grant No. T45-701/22-R) and Shenzhen Science and Technology Innovation Commission (SGDX20220530111405040). Part of the described research work is conducted in the JC STEM Lab of Robotics for Soft Materials funded by The Hong Kong Jockey Club Charities Trust.

{
    \small
    \bibliographystyle{ieeenat_fullname}
    \bibliography{main}
}

\clearpage
\setcounter{page}{1}
\setcounter{section}{0}
\setcounter{figure}{0}
\setcounter{table}{0}
\maketitlesupplementary

\renewcommand{\thesection}{S\arabic{section}}
\renewcommand{\thesubsection}{S\arabic{subsection}}
\renewcommand{\thetable}{S\arabic{table}}
\renewcommand{\thefigure}{S\arabic{figure}}

\centerline{\large{\textbf{Outline}}}
\vspace{0.3cm}

In this supplementary file, we provide additional applications and potential usages of \textit{DRK}, an introduction to the DiverseScenes dataset, implementation details, and further results that could not be included in the main paper due to space constraints. The content is organized as follows:

\begin{itemize}
    \item Sec.~\ref{supsec:mesh2drk}: More results of seamless conversion from mesh models to \textit{DRK} representations, bridging millions of 3D assets with high-fidelity reconstructed scenes.
    \item Sec.~\ref{supsec:diversescenes}: Introduction to the DiverseScenes dataset.
    \item Sec.~\ref{supsec:experiments}: Additional experimental results on public datasets and an analysis of limitations.
    \item Sec.~\ref{supsec:method_details}: Detailed implementation of the \textit{DRK} framework.
\end{itemize}

\section{Converting Mesh to DRK}
\label{supsec:mesh2drk}

\begin{figure}[thb]
\begin{center}
\vspace{-2mm}
\includegraphics[width=\linewidth]{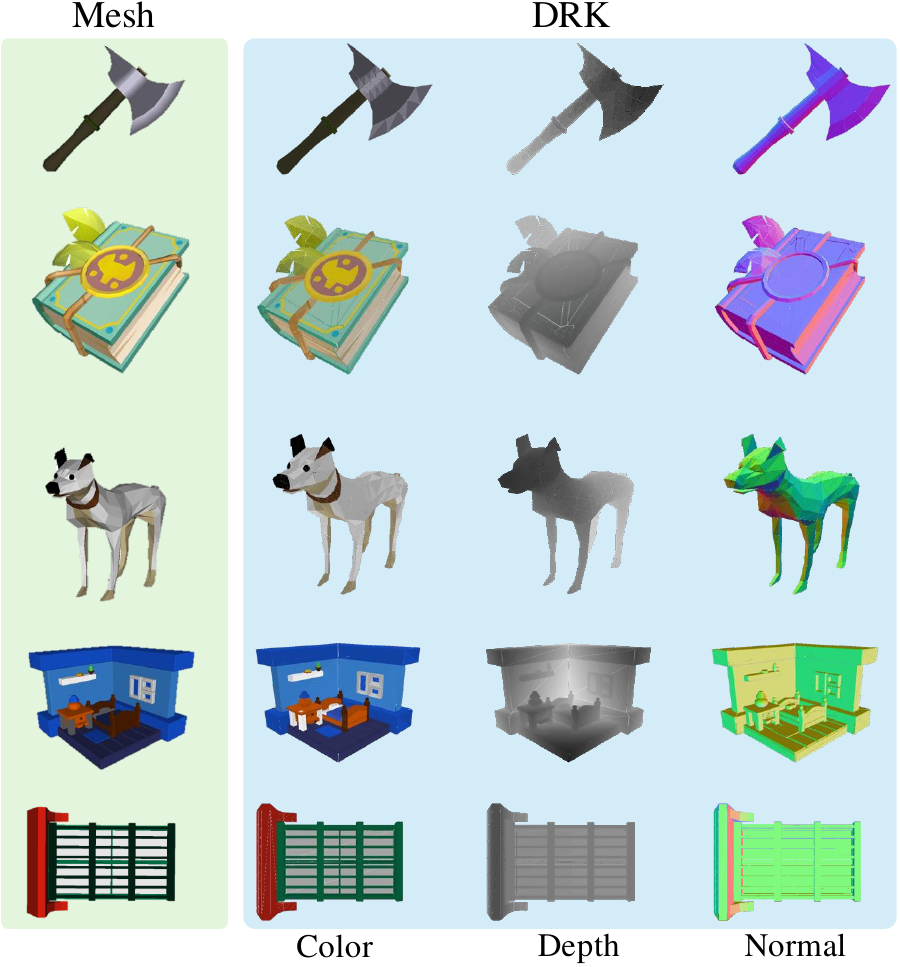}
\vspace{-10mm}
\end{center}
\caption{Examples of converting various 3D assets. The normal of \textit{DRK} can be used for shading under illumination. UV texture mapping is applicable for \textit{DRK} in future implementations.}
\label{fig:mesh2drk}
\vspace{-3mm}
\end{figure}

We present more examples of converting 3D mesh assets to \textit{DRK} within seconds, without the need for training data preparation or optimization, in Fig.~\ref{fig:mesh2drk}. Rendered depth and normal images are also provided. In these examples, the \textit{DRK} is kernel-wise colored and shaded over the base color using the normal and predefined illumination. 
In the future, by assigning UV attributes to \textit{DRK} and rendering them into images, material properties such as albedo, roughness, and metallicity can be retrieved from the material maps, enabling deferred rendering using the rendered normals.
This capability allows \textit{DRK} to handle traditional assets and compose scenes reconstructed from real-world multi-views and man-made artistic 3D assets.

It is worth noting that cache-sorting has minimal impact on reconstruction quality but plays a crucial role in the conversion from mesh to \textit{DRK}. This is because Mesh2\textit{DRK} produces a compact geometric representation, where each \textit{DRK} kernel represents a relatively larger unit of a mesh face compared to those learned from multi-view images, which use smaller units to capture high-frequency appearance details. Cache-sorting ensures that the sorting order is nearly correct, resulting in satisfactory conversion outcomes.

\section{DiverseScenes Dataset}
\label{supsec:diversescenes}

\begin{figure}[thb]
\begin{center}
\vspace{-2mm}
\includegraphics[width=\linewidth]{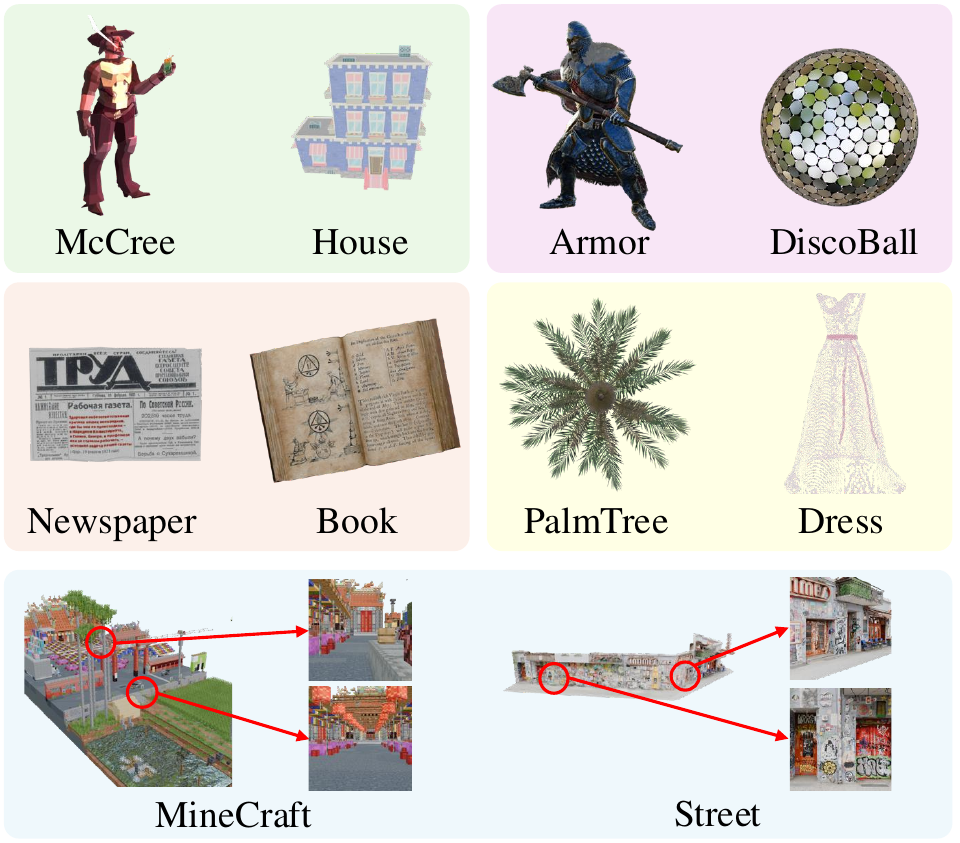}
\vspace{-10mm}
\end{center}
\caption{Overview of DiverseScenes. The dataset is composed of five categories: simple, specular, detailed geometry, fine texture, and large scale.}
\label{fig:diverse}
\vspace{-3mm}
\end{figure}

We collected 10 scenes from Sketchfab\footnote{https://sketchfab.com/}, encompassing a variety of 3D scene types. As illustrated in Fig.~\ref{fig:diverse}, the dataset includes scenes with simple geometry and textures (e.g., McCree, House), detailed textures (e.g., Newspaper, Book), fine geometry (e.g., PalmTree, Dress), and large scales (e.g., Minecraft, Street). The training set consists of 200 views, and the test set includes 30 views, sampled from the unit sphere. For Minecraft and Street, there are 230 training views, with the train and test views simulating a walking camera along specified paths.

\begin{table}[ht]
  \centering
    \adjustbox{width={\linewidth},keepaspectratio}{
    \begin{tabular}{lSSSSSS}
        \toprule[1pt]
        \multirow{2}{*}{Methods} &
          \multicolumn{3}{c}{\textbf{McCree}} &
          \multicolumn{3}{c}{\textbf{House}} \\
          \cmidrule(lr){2-4}
          \cmidrule(lr){5-7}
          & {\textbf{PSNR($\uparrow$)}} & {\textbf{LPIPS($\downarrow$)}} & {\textbf{SSIM($\uparrow$)}} 
          & {\textbf{PSNR($\uparrow$)}} & {\textbf{LPIPS($\downarrow$)}} & {\textbf{SSIM($\uparrow$)}} \\
          \midrule
            2D-GS
            & \cellcolor{second}{40.95} & \cellcolor{third}{.0067} & \cellcolor{second}{.9985}
            & \cellcolor{third}{43.25} & \cellcolor{third}{.0351} & \cellcolor{third}{.9953} \\
            3D-GS
            & {39.15} & {.0115} & {.9975}
            & {39.87} & {.0507} & {.9907} \\
            3D-HGS  
            & {40.64} & \cellcolor{second}{.0062} & {.9965}
            & {42.17} & {.0335} & {.9897} \\
            GES
            & {40.48} & {.0086} & {.9982}
            & {41.42} & {.0446} & {.9927} \\
          \cmidrule(lr){2-4}
          \cmidrule(lr){5-7}
            Ours (S2)
            & {39.76} & {.0104} & {.9977}
            & {40.92} & {.0462} & {.9914} \\
            Ours (S1)
            & \cellcolor{third}{40.90} & \cellcolor{third}{.0067} & \cellcolor{second}{.9985}
            & \cellcolor{second}{43.57} & \cellcolor{second}{.0269} & \cellcolor{second}{.9963} \\
            Ours
            & \cellcolor{best}{41.70} & \cellcolor{best}{.0047} & \cellcolor{best}{.9989}
            & \cellcolor{best}{45.22} & \cellcolor{best}{.0163} & \cellcolor{best}{.9980} \\
        \midrule
        \multirow{2}{*}{Methods} &
          \multicolumn{3}{c}{\textbf{Book}} &
          \multicolumn{3}{c}{\textbf{Newspaper}} \\
          \cmidrule(lr){2-4}
          \cmidrule(lr){5-7}
          & {\textbf{PSNR($\uparrow$)}} & {\textbf{LPIPS($\downarrow$)}} & {\textbf{SSIM($\uparrow$)}}
          & {\textbf{PSNR($\uparrow$)}} & {\textbf{LPIPS($\downarrow$)}} & {\textbf{SSIM($\uparrow$)}} \\
          \midrule
            2D-GS
            & {38.20} & {.1199} & {.9673}
            & {46.97} & {.0048} & \cellcolor{best}{.9995} \\
            3D-GS
            & {42.25} & {.1211} & {.9944}
            & {46.89} & {.0045} & {.9991} \\
            3D-HGS
            & {43.48} & \cellcolor{best}{.0775} & {.9842}
            & \cellcolor{third}{48.89} & \cellcolor{best}{.0019} & {.9981} \\
            GES
            & \cellcolor{third}{43.49} & {.1122} & {.9960}
            & {46.97} & {.0036} & \cellcolor{second}{.9993} \\
          \cmidrule(lr){2-4}
          \cmidrule(lr){5-7}
            Ours (S2)
            & {43.28} & {.1042} & \cellcolor{third}{.9961}
            & {47.75} & {.0032} & {.9991} \\
            Ours (S1)
            & \cellcolor{second}{45.57} & \cellcolor{third}{.0904} & \cellcolor{second}{.9976}
            & \cellcolor{second}{49.76} & \cellcolor{third}{.0023} & {.9992} \\
            Ours
            & \cellcolor{best}{47.28} & \cellcolor{second}{.0801} & \cellcolor{best}{.9983}
            & \cellcolor{best}{50.88} & \cellcolor{best}{.0019} & \cellcolor{second}{.9993} \\
        \midrule
        \multirow{2}{*}{Methods} &
          \multicolumn{3}{c}{\textbf{Dress}} &
          \multicolumn{3}{c}{\textbf{PalmTree}} \\
          \cmidrule(lr){2-4}
          \cmidrule(lr){5-7}
          & {\textbf{PSNR($\uparrow$)}} & {\textbf{LPIPS($\downarrow$)}} & {\textbf{SSIM($\uparrow$)}}
          & {\textbf{PSNR($\uparrow$)}} & {\textbf{LPIPS($\downarrow$)}} & {\textbf{SSIM($\uparrow$)}} \\
          \midrule
            2D-GS
            & {26.04} & {.0367} & {.9807}
            & {33.76} & {.0659} & {.9593} \\
            3D-GS
            & {25.45} & {.0378} & {.9782}
            & {32.82} & {.0725} & {.9533} \\
            3D-HGS
            & {27.30} & \cellcolor{second}{.0217} & {.9773}
            & {33.54} & {.0595} & {.9423} \\
            GES
            & {27.31} & {.0254} & \cellcolor{third}{.9859}
            & {33.41} & {.0682} & {.9549} \\
          \cmidrule(lr){2-4}
          \cmidrule(lr){5-7}
            Ours (S2)
            & \cellcolor{third}{27.56} & {.0362} & {.9770}
            & \cellcolor{third}{33.80} & \cellcolor{third}{.0592} & \cellcolor{third}{.9648} \\
            Ours (S1)
            & \cellcolor{second}{29.19} & \cellcolor{third}{.0225} & \cellcolor{second}{.9899}
            & \cellcolor{second}{34.14} & \cellcolor{second}{.0562} & \cellcolor{second}{.9677} \\
            Ours
            & \cellcolor{best}{29.61} & \cellcolor{best}{.0196} & \cellcolor{best}{.9917}
            & \cellcolor{best}{34.25} & \cellcolor{best}{.0536} & \cellcolor{best}{.9688} \\
        \midrule
        \multirow{2}{*}{Methods} &
          \multicolumn{3}{c}{\textbf{Armor}} &
          \multicolumn{3}{c}{\textbf{Discoball}} \\
          \cmidrule(lr){2-4}
          \cmidrule(lr){5-7}
          & {\textbf{PSNR($\uparrow$)}} & {\textbf{LPIPS($\downarrow$)}} & {\textbf{SSIM($\uparrow$)}}
          & {\textbf{PSNR($\uparrow$)}} & {\textbf{LPIPS($\downarrow$)}} & {\textbf{SSIM($\uparrow$)}} \\
          \midrule
            2D-GS
            & \cellcolor{third}{32.65} & \cellcolor{third}{.0500} & \cellcolor{third}{.9742}
            & {21.70} & \cellcolor{second}{.0639} & {.9357} \\
            3D-GS
            & {31.51} & {.0687} & {.9684}
            & \cellcolor{best}{22.81} & {.0828} & {.9361} \\
            3D-HGS
            & {32.26} & {.0518} & {.9512}
            & \cellcolor{third}{22.28} & \cellcolor{best}{.0505} & {.9381} \\
            GES
            & {31.98} & {.0603} & {.9713}
            & {22.09} & \cellcolor{third}{.0661} & {.9386} \\
          \cmidrule(lr){2-4}
          \cmidrule(lr){5-7}
            Ours (S2)
            & {32.01} & {.0592} & {.9736}
            & {22.13} & {.0816} & \cellcolor{third}{.9404} \\
            Ours (S1)
            & \cellcolor{second}{33.27} & \cellcolor{second}{.0430} & \cellcolor{second}{.9791}
            & {22.26} & {.0832} & \cellcolor{second}{.9414} \\
            Ours
            & \cellcolor{best}{34.07} & \cellcolor{best}{.0346} & \cellcolor{best}{.9814}
            & \cellcolor{second}{22.31} & {.0838} & \cellcolor{best}{.9415} \\
        \midrule
        \multirow{2}{*}{Methods} &
          \multicolumn{3}{c}{\textbf{Minecraft}} &
          \multicolumn{3}{c}{\textbf{Street}} \\
          \cmidrule(lr){2-4}
          \cmidrule(lr){5-7}
          & {\textbf{PSNR($\uparrow$)}} & {\textbf{LPIPS($\downarrow$)}} & {\textbf{SSIM($\uparrow$)}}
          & {\textbf{PSNR($\uparrow$)}} & {\textbf{LPIPS($\downarrow$)}} & {\textbf{SSIM($\uparrow$)}} \\
          \midrule
            2D-GS
            & {23.15} & {.4119} & {.7262}
            & {32.45} & {.0881} & {.9514} \\
            3D-GS
            & {25.81} & {.3516} & \cellcolor{third}{.8116}
            & {38.56} & {.0591} & {.9917} \\
            3D-HGS
            & \cellcolor{second}{26.10} & \cellcolor{third}{.3070} & {.7568}
            & \cellcolor{third}{40.23} & \cellcolor{second}{.0274} & {.9873} \\
            GES
            & {25.83} & {.3554} & {.8072}
            & {37.51} & {.0594} & {.9902} \\
          \cmidrule(lr){2-4}
          \cmidrule(lr){5-7}
            Ours (S2)
            & {25.01} & {.3584} & {.8019}
            & {38.12} & {.0596} & \cellcolor{third}{.9950} \\
            Ours (S1)
            & \cellcolor{third}{25.98} & \cellcolor{second}{.3057} & \cellcolor{second}{.8410}
            & \cellcolor{second}{41.58} & \cellcolor{third}{.0312} & \cellcolor{second}{.9981} \\
            Ours
            & \cellcolor{best}{26.84} & \cellcolor{best}{.2512} & \cellcolor{best}{.8755}
            & \cellcolor{best}{43.72} & \cellcolor{best}{.0183} & \cellcolor{best}{.9989} \\
        \midrule
        \multirow{2}{*}{Methods} &
          \multicolumn{6}{c}{\textbf{Average}}\\
          \cmidrule(lr){2-7}
          & {\textbf{PSNR($\uparrow$)}} & {\textbf{LPIPS($\downarrow$)}} & {\textbf{SSIM($\uparrow$)}}
          & {\textbf{Num($\downarrow$)}} & {\textbf{Size($\downarrow$)}} & {\textbf{FPS($\uparrow$)}} \\
          \midrule
            2D-GS
            & {33.92} & {.0881} & {.9514}
            & {359K} & {83.6M} & \cellcolor{best}{251.3} \\
            3D-GS
            & {34.41} & {.0861} & {.9621}
            & {346K} & {82.0M} & \cellcolor{second}{247.1} \\
            3D-HGS  
            & \cellcolor{third}{35.68} & \cellcolor{second}{.0637} & {.9521}
            & {373K} & {89.6M} & {154.5} \\
            GES
            & {35.05} & {.0804} & {.9634}
            & {330K} & {78.1M} & {227.4} \\
          \cmidrule(lr){2-4}
          \cmidrule(lr){5-7}
            Ours (S2)
            & {35.03} & {.0823} & \cellcolor{third}{.9637}
            & \cellcolor{best}{42K} & \cellcolor{best}{12.3M} & \cellcolor{third}{234.9} \\
            Ours (S1)
            & \cellcolor{second}{36.62} & \cellcolor{third}{.0668} & \cellcolor{second}{.9701}
            & \cellcolor{second}{109K} & \cellcolor{second}{32.1M} & {119.2} \\
            Ours
            & \cellcolor{best}{37.58} & \cellcolor{best}{.0564} & \cellcolor{best}{.9752}
            & \cellcolor{third}{260K} & \cellcolor{third}{76.6M} & {77.5} \\
        \bottomrule[1pt]
      \end{tabular}}
  \vspace{-2mm}
  \caption{We show the PSNR, LPIPS, and SSIM metrics for novel view synthesis on DiverseScenes.}\label{tab:sup_diversescenes}
  \vspace{-2mm}
\end{table}

To offer a more comprehensive overview of the DiverseScenes Dataset, we manually annotated the attributes of each scene, as summarized in Table~\ref{tab:diverse}. Note that some attributes overlap; for instance, both Armor and Street include fine textures.
To clarify the performance of methods on DiverseScenes, we present the per-scene results in Table~\ref{tab:sup_diversescenes}. These results are also summarized by scene categories in the main paper.

\begin{table}[ht]
\centering
    \adjustbox{width={\linewidth},keepaspectratio}{
\begin{tabular}{l|SS|SS|SS|SS}
 \hline
 \multirow{2}{*}{Scene} & \multicolumn{2}{c|}{Geometry} & \multicolumn{2}{c|}{Texture} & \multicolumn{2}{c|}{Scale} & \multicolumn{2}{c}{Material} \\
 \cline{2-9}
 & {Coarse} & {Fine} & {Coarse} & {Fine} & {Small} & {Large} & {Diffuse} & {Specular} \\
 \hline \hline
 McCree & \checkmark & \texttimes & \checkmark & \texttimes & \checkmark & \texttimes & \checkmark & \texttimes  \\
 \hline
 House & \checkmark & \texttimes & \checkmark & \texttimes & \checkmark & \texttimes & \checkmark & \texttimes  \\
 \hline \hline
 Book & \checkmark & \texttimes & \texttimes & \cellcolor{lightgray}\checkmark & \checkmark & \texttimes & \checkmark & \texttimes  \\
 \hline
 Newspaper & \checkmark & \texttimes & \texttimes & \cellcolor{lightgray}\checkmark & \checkmark & \texttimes & \checkmark & \texttimes  \\
 \hline \hline
 Dress & \texttimes & \cellcolor{lightgray}\checkmark & \checkmark & \texttimes & \checkmark & \texttimes & \checkmark & \texttimes  \\ \hline 
 PalmTree & \texttimes & \cellcolor{lightgray}\checkmark & \checkmark & \texttimes & \checkmark & \texttimes & \checkmark & \texttimes  \\
 \hline \hline
 Armor & \checkmark & \texttimes & \texttimes & \cellcolor{lightgray}\checkmark & \checkmark & \texttimes & \texttimes & \cellcolor{lightgray}\checkmark  \\ \hline
 DiscoBall & \checkmark & \texttimes & \checkmark & \texttimes & \checkmark & \texttimes & \texttimes & \cellcolor{lightgray}\checkmark  \\ \hline \hline
 Street & \checkmark & \texttimes & \texttimes & \cellcolor{lightgray}\checkmark & \texttimes & \cellcolor{lightgray}\checkmark & \checkmark & \texttimes  \\
 \hline
 MineCraft & \checkmark & \texttimes & \checkmark & \texttimes & \texttimes & \cellcolor{lightgray}\checkmark & \checkmark & \cellcolor{lightgray}\checkmark  \\
 \hline
\end{tabular}
}
\vspace{-3mm}
\caption{Summary of scene attribute annotations.}
\label{tab:diverse}
\vspace{-2mm}
\end{table}

\section{More Experiments \& Limitation Analysis}
\label{supsec:experiments}

\paragraph{Evaluation on NeRF-Synthetic Scenes.}

We conducted a quantitative evaluation on NeRF-Synthetic~\cite{mildenhall2021nerf} scenes, reporting PSNR scores, primitive numbers, and model sizes in Table~\ref{tab:sup-nerf-transposed}. Unlike DiverseScenes, the training cameras for NeRF-Synthetic scenes (except Ficus) are sampled from the upper hemisphere, and some test views fall outside the range covered by the training views. This setup partially assesses performance on view extrapolation. Our \textit{DRK} method demonstrates superior performance. While the kernel number for \textit{DRK} is slightly larger than for 3D-GS~\cite{kerbl2023d}, \textit{DRK} (S1) maintains a compact number and outperforms other methods in PSNR scores. \textit{DRK} (S3) has a very small kernel number and model size, with an average PSNR still comparable to others.

\begin{table}[t]
  \centering
    \adjustbox{width={\linewidth},keepaspectratio}{
    \begin{tabular}{lSSSS|SSS}
        \toprule[1pt]
        Scene & {2D-GS} & {3D-GS} & {3D-HGS} & {GES} & {\textit{DRK} (S2)} & {\textit{DRK} (S1)} & {\textit{DRK}} \\
        \cmidrule(r){1-8}
        Chair 
        & {34.88} & \cellcolor{best}{35.83} & {34.29} & {34.05} & {34.38} & \cellcolor{third}{35.28} & \cellcolor{second}{35.61} \\
        Drums 
        & {25.67} & \cellcolor{second}{26.15} & \cellcolor{best}{26.29} & {26.05} & {25.90} & {26.12} & \cellcolor{third}{26.13} \\
        Ficus 
        & \cellcolor{third}{35.80} & {34.87} & {35.45} & {35.27} & {35.56} & \cellcolor{second}{36.27} & \cellcolor{best}{36.50} \\
        Hotdog 
        & {36.89} & \cellcolor{third}{37.72} & {37.54} & {37.13} & {37.19} & \cellcolor{second}{37.84} & \cellcolor{best}{38.17} \\
        Lego 
        & {34.82} & \cellcolor{second}{35.78} & {33.92} & {33.73} & {33.90} & \cellcolor{third}{35.38} & \cellcolor{best}{36.25} \\
        Materials 
        & \cellcolor{second}{30.14} & {30.00} & {29.88} & {29.74} & {29.38} & \cellcolor{second}{30.14} & \cellcolor{best}{30.48} \\
        Mic 
        & {34.38} & {35.36} & \cellcolor{best}{36.58} & \cellcolor{third}{35.73} & {35.17} & {35.70} & \cellcolor{second}{36.00} \\
        Ship 
        & {31.09} & {30.80} & \cellcolor{third}{31.10} & {30.94} & {30.84} & \cellcolor{second}{31.28} & \cellcolor{best}{31.42} \\
        \cmidrule(r){1-8}
        Avg PSNR
        & {32.96} & \cellcolor{third}{33.32} & {33.13} & {32.83} & {32.79} & \cellcolor{second}{33.50} & \cellcolor{best}{33.82} \\
        \cmidrule(r){1-8}
        Num 
        & {107K} & {131K} & {83K} & \cellcolor{second}{73K} & \cellcolor{best}{32K} & \cellcolor{third}{75K} & {158K} \\
        Size 
        & {25.0M} & {31.1M} & \cellcolor{third}{20.0M} & \cellcolor{second}{17.4M} & \cellcolor{best}{9.6M} & {22.0M} & {46.6M} \\
        \bottomrule[1pt]
      \end{tabular}
    }
\vspace{-1mm}
  \caption{PSNR scores, primitive numbers, and model sizes on NeRF-Synthetic~\cite{mildenhall2021nerf} scenes (transposed).}\label{tab:sup-nerf-transposed}
  \vspace{-4mm}
\end{table}

\paragraph{Ablation Study}
To gain a comprehensive understanding of DRK, we perform ablation studies on its attributes to assess their impact. We remove the effects of sharpening and curvature learning by setting \(\tau = 0\) and \(\eta = 0\), respectively. We also examine the impact of the hyper-parameter \(K\). Ablation results on DiverseScenes are shown in Tab.~\ref{tab:ablation}. We found that \(K\) significantly influences performance, with \(K = 3\) causing the greatest drop. Additionally, \(\eta\) is more vital than \(\tau\), though both enhance representation capability.

\begin{table}[t]
  \centering
  \setlength\tabcolsep{10pt}
  \resizebox{0.95\linewidth}{!}{
    \begin{tabular}{@{\extracolsep{\fill}} lSSSSSS }
        \toprule[1pt]
        \multirow{1}{*}{Methods} &
          \multicolumn{1}{c}{\textbf{$\tau=0$}} &
          \multicolumn{1}{c}{\textbf{$\eta=0$}} &
          \multicolumn{1}{c}{\textbf{$K=3$}} &
          \multicolumn{1}{c}{\textbf{$K=5$}} &
          \multicolumn{1}{c}{\textbf{$K=8$}} \\
          \midrule
            \textit{DRK} (S2)
            & \cellcolor{second}{34.97}
            & \cellcolor{third}{34.82}
            & {33.93}
            & {34.47}
            & \cellcolor{best}{35.03} \\
            \textit{DRK} (S1)
            & \cellcolor{second}{36.22}
            & \cellcolor{third}{36.04}
            & {35.50}
            & {35.88}
            & \cellcolor{best}{36.62} \\
            \textit{DRK}
            & \cellcolor{second}{37.46}
            & \cellcolor{third}{37.27}
            & {36.60}
            & {36.85}
            & \cellcolor{best}{37.58} \\
            
        \bottomrule[1pt]
      \end{tabular}
      }
  \caption{Ablation study on the impact of $\tau$, $\eta$, and $K$.}\label{tab:ablation}
\end{table}

\paragraph{Evaluation on Tank\&Temple Scenes.}

To assess our method's performance on more challenging scenes with imperfect camera conditions due to dynamic objects and changing exposures, we evaluated the Tank\&Temple~\cite{Knapitsch2017} datasets. We used 9 scenes in total, including 8 intermediate scenes and the Truck scene. We report the PSNR scores, foreground-only PSNR scores (M-PSNR), primitive numbers, and model sizes in Table~\ref{tab:sup-tt}. The results indicate that \textit{DRK} faces significant challenges with this dataset, likely due to higher camera error estimated by COLMAP~\cite{schoenberger2016sfm}. The Tank\&Temple datasets are captured in dynamic environments with moving pedestrians and changing exposure, making camera estimation more difficult than in MipNeRF-360~\cite{barron2022mip}, where objects are primarily diffuse and free from view-dependent effects, transients, or significant sunlight exposure changes. To further investigate the robustness of \textit{DRK}, we conducted evaluations with noisy camera data.

\begin{table}[h]
  \centering
    \adjustbox{width={\linewidth},keepaspectratio}{
    \begin{tabular}{lSSSS|SSS }
        \toprule[1pt]
        Methods
          & {2D-GS} & {3D-GS} & {3D-HGS} & {GES} & {Ours (S2)} & {Ours (S1)} & {Ours} \\
          \cmidrule(r){1-8} %
            \textbf{PSNR}
            & \cellcolor{third}{20.65} & \cellcolor{second}{21.09} & \cellcolor{best}{21.59} & {20.58} & {20.20} & {20.31} & {20.41} \\
            \textbf{M-PSNR}
            & {26.50} & \cellcolor{second}{26.92} & \cellcolor{best}{27.58} & \cellcolor{third}{26.56} & {26.36} & {26.41} & {26.37} \\
            \textbf{Num}
            & {1168K} & {275K} & {267K} & \cellcolor{third}{259K} & \cellcolor{best}{173K} & \cellcolor{second}{212K} & {383K} \\
            \textbf{Size}
            & {271.9M} & {65.4M} & {64.2M} & \cellcolor{second}{61.7M} & \cellcolor{best}{50.9M} & \cellcolor{third}{62.4M} & {112.6M} \\
        \bottomrule[1pt]
      \end{tabular}
      }
\vspace{-3mm}
  \caption{Quantitative evaluation on Tank\&Temple scenes.}\label{tab:sup-tt}
  \vspace{-1mm}
\end{table}

\paragraph{Robustness against Camera Noise}

To evaluate the performance of \textit{DRK} under varying levels of camera noise, we simulated camera noise with different standard deviations (Std). The PSNR scores on DiverseScenes with noisy cameras are reported in Table~\ref{tab:noise_ablation}. We observed that the PSNR scores of \textit{DRK} drop significantly as the camera noise increases, whereas the performance of 3D-GS degrades more smoothly and slightly.

\begin{table}[h]
  \centering
    \adjustbox{width={.8\linewidth},keepaspectratio}{
    \begin{tabular}{l|SSSS }
        \toprule[1pt]
        \multirow{1}{*}{Noise \textit{Std}} &
          \multicolumn{1}{c}{{3D-GS}} &
          \multicolumn{1}{c}{{\textit{DRK}}} &
          \multicolumn{1}{c}{{\textit{DRK} (S1)}} &
          \multicolumn{1}{c}{{\textit{DRK} (S2)}} \\
          \midrule
            \textit{$0$}
            & {34.41}
            & \cellcolor{best}{37.58}
            & \cellcolor{second}{36.62}
            & \cellcolor{third}{35.03} \\
            \textit{$1e-3$}
            & \cellcolor{best}{33.44}
            & \cellcolor{second}{31.59}
            & \cellcolor{third}{31.19}
            & {30.88} \\
            \textit{$2.5e-3$}
            & \cellcolor{best}{31.32}
            & \cellcolor{second}{29.27}
            & \cellcolor{third}{28.89}
            & {28.60} \\
            \textit{$5e-3$}
            & \cellcolor{best}{29.37}
            & \cellcolor{second}{27.85}
            & \cellcolor{third}{27.44}
            & {27.14} \\
            
        \bottomrule[1pt]
      \end{tabular}
      }
  \vspace{-2mm}
  \caption{Average PSNR scores on DiverseScenes of 3D-GS and \textit{DRK} under different levels of camera noise.}\label{tab:noise_ablation}
  \vspace{-4mm}
\end{table}

\begin{figure}[hb]
\begin{center}
\vspace{-2mm}
\includegraphics[width=\linewidth]{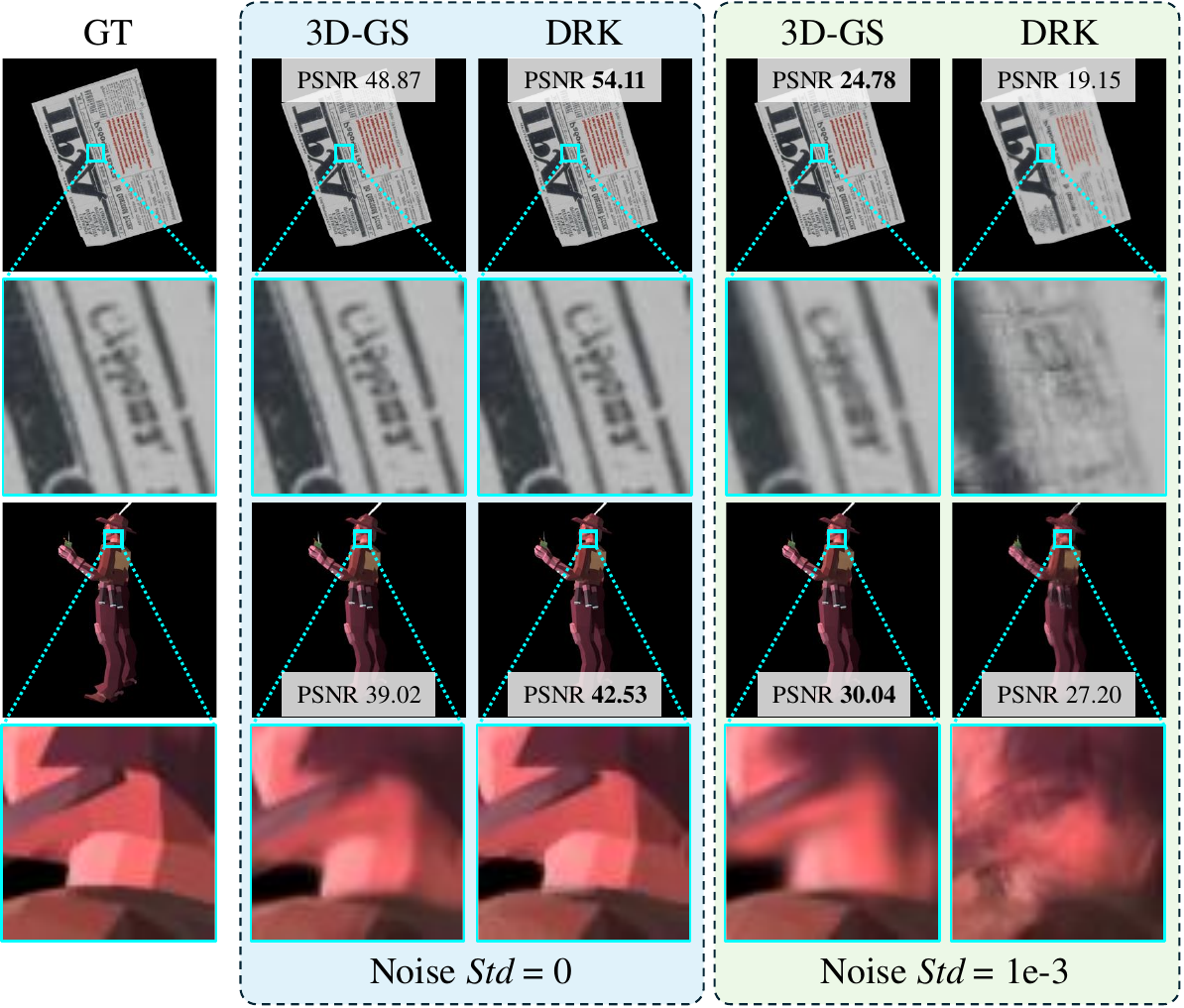}
\vspace{-10mm}
\end{center}
\caption{Rendering results of 3D-GS and \textit{DRK} trained on cameras with and without noise.}
\label{fig:robustness_camera_noise}
\vspace{-3mm}
\end{figure}

Fig.~\ref{fig:robustness_camera_noise} shows the rendering results of 3D-GS and \textit{DRK} trained with both accurate and noisy cameras. When trained with accurate cameras, \textit{DRK} achieves higher-quality rendering with sharper and clearer appearances. However, even with very small camera noise, the performance of \textit{DRK} deteriorates significantly, producing blurrier and more chaotic results compared to 3D-GS. In contrast, 3D-GS maintains the ability to model the coarse appearance of the scene under noisy conditions. 
These results demonstrate that \textit{DRK} is less robust to camera noise, which may explain its performance drop on the Tank\&Temple dataset.

\begin{figure*}[thb]
\begin{center}
\includegraphics[width=.8\linewidth]{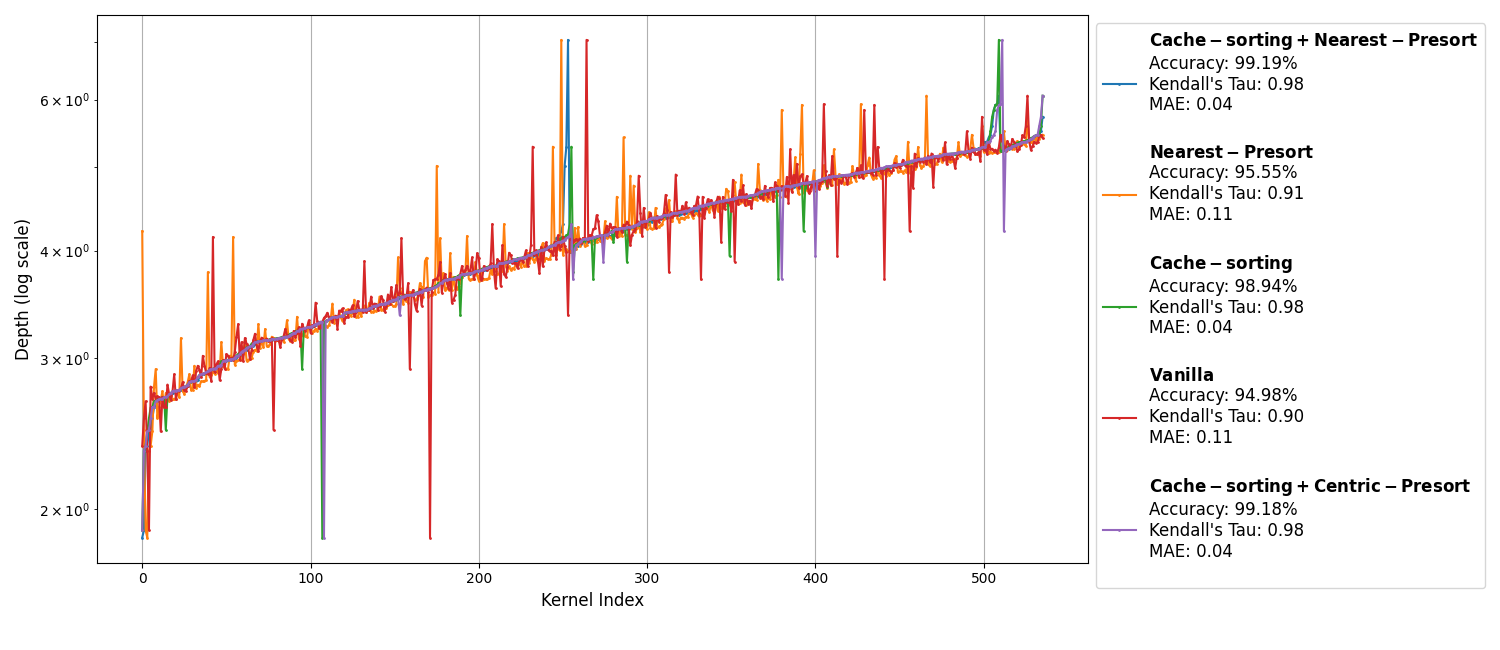}
\vspace{-9mm}
\end{center}
\caption{Sorting accuracy comparisons: We found that pre-sorting tile-kernel pairs based on the nearest distance, combined with cache-sorting, achieves the highest accuracy. Cache-sorting is sufficiently effective in correcting most sorting disorders.}
\label{fig:sup-cache-sort}
\vspace{-3mm}
\end{figure*}

\begin{figure}[thb]
\begin{center}
\vspace{-2mm}
\includegraphics[width=.7\linewidth]{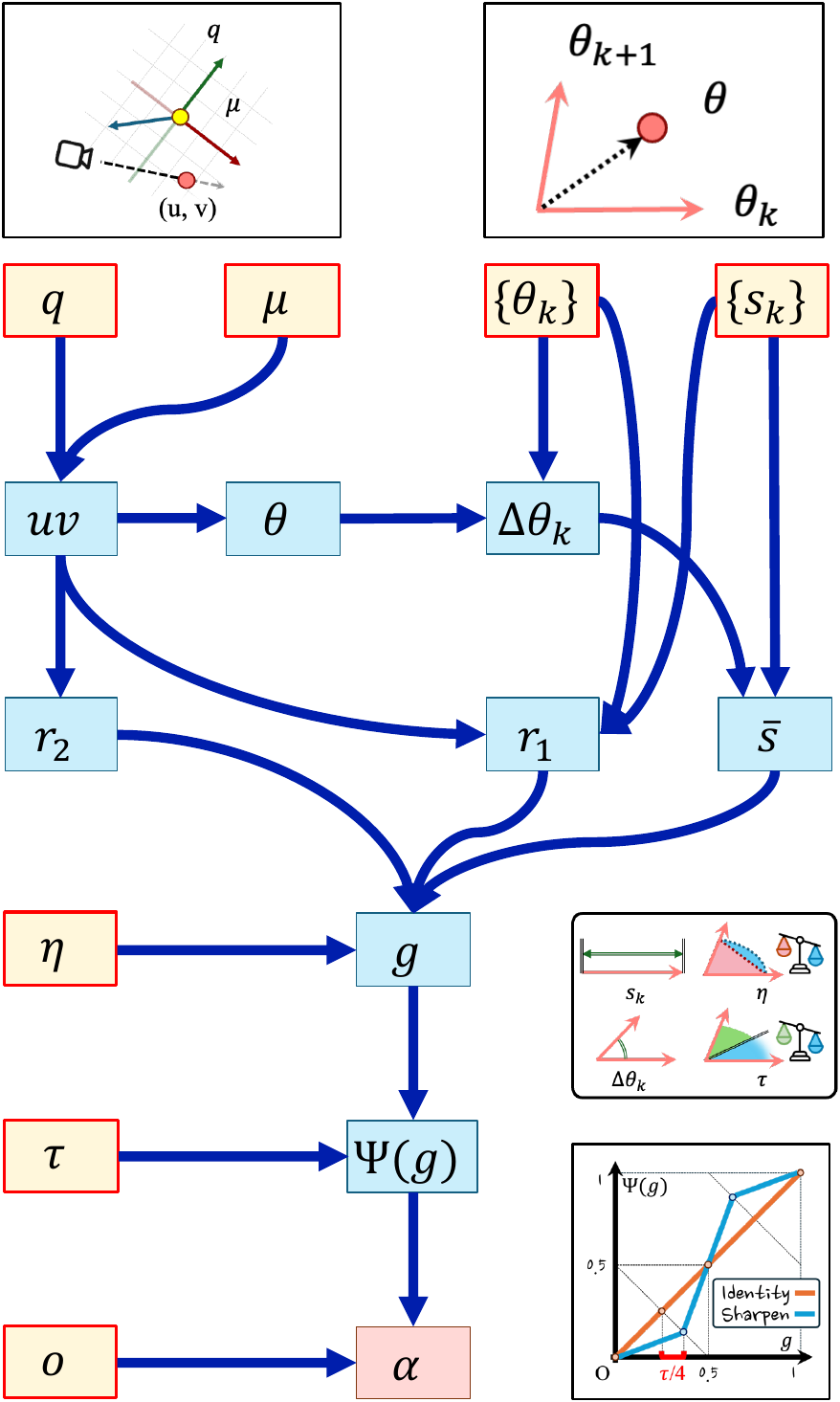}
\vspace{-5mm}
\end{center}
\caption{The "Tensor Graph" of \textit{DRK}, showing the dependence between parameters to optimize, intermediate variables, and the final output ($\alpha$).}
\label{fig:tensorflow}
\vspace{-3mm}
\end{figure}

\section{Method Details}
\label{supsec:method_details}

\paragraph{Parametrization.}
\textit{DRK} attributes are modeled using unconstrained learnable parameters in $(-\infty, \infty)$, with appropriate activation functions to ensure valid ranges. We apply a sigmoid function to constrain opacity $o$ to $(0, 1)$, an exponential function for scale activation $s_k$, and a normalization function for rotation quaternion $q$. The sharpness $\tau$ is bounded within $(-0.1, 0.99)$ through a composite function combining sigmoid and linear remapping, while the {$L1\&L2$} blending weight $\eta$ is activated using a sigmoid function. For the basis angle $\theta_k$, we employ a three-step activation process: first applying a sigmoid function, then adding a residual term $\frac{1}{K-2}$ to maintain minimum angular separation, applying cumulative summation to enforce monotonic increase, and finally normalizing to make $s_K=2\pi$. The residual term prevents basis polar angles from exceeding $\pi$, thereby avoiding degradation in the representation. {We set $K=8$ to balance flexibility and memory efficiency.}

\paragraph{Model Training.}
Following 3D-GS, we optimize model parameters and dynamically adjust kernel density through an adaptive training process. {For \emph{DRK-specific} parameters} - sharpness $\tau$, blending weight $\eta$, basis angles $\theta_k$, and scales $s_k$ - we set a uniform learning rate of $5e^{-3}$ and decay them gradually to the rate $1e^{-2}$$\times$ at the end of training (35K steps). We implement three density control configurations through the 2D screen gradient densification threshold and opacity pruning threshold pairs: ($5e^{-4}$, $5e^{-2}$) for density comparable to 3D-GS, ($1e^{-3}$, $5e^{-2}$) for Sparse Level 1 (\textbf{S1}), and ($2e^{-3}$, $1e^{-1}$) for Sparse Level 2 (\textbf{S2}).

\paragraph{Tensor Graph of \textit{DRK}}

To provide a detailed overview of \textit{DRK}, we present its "Tensor Graph," which illustrates the flow from the learnable leaf parameters through the intermediate variables, ultimately leading to the outputs. The graph is shown in Fig.~\ref{fig:tensorflow}. Blue arrows represent data dependencies, along which gradients are back-propagated in reverse during optimization.

\paragraph{Cache-Sorting}

To clarify the cache-sorting algorithm, we briefly summarize the process in Algorithm~\ref{alg:cache-sort}. As discussed in StopThePop~\cite{radl2024stopthepop}, the backward processing must also be adjusted to proceed from front to back to maintain consistency with forward rendering.

\begin{algorithm}
\caption{Cache-Sorting}
\label{alg:cache-sort}
\KwIn{A cache chain with limited size, a new \textit{DRK} (with an index and depth)}
\KwOut{The index of textit{DRK} or a status code}

\If{the cache is empty}{
    \If{the new \textit{DRK} is invalid}{
        \Return a finish code\;
    }
    Initialize the cache with the new \textit{DRK}\;
    \Return success\;
}

\If{the new \textit{DRK} is invalid}{
    Pop the \textit{DRK} from the head\;
    \Return the index of the popped \textit{DRK}\;
}

\If{the cache is full}{
    Mark the head \textit{DRK} to be popped\;
}

Determine where to insert the new \textit{DRK} by scanning the cache, guided by the \textit{DRK}'s depth\;
Adjust the pointers in the cache to insert the new \textit{DRK} at the correct position\;

\If{the cache was full}{
    Pop the oldest \textit{DRK}\;
}

\Return the index of the popped \textit{DRK} (or success if none was popped)\;

\end{algorithm}

We evaluate the effectiveness of cache-sorting using a cache length of 8. \textit{DRK} kernels are randomly sampled from the space, and the depths of \textit{DRK} intersections processed in a front-to-back order are visualized in Fig.~\ref{fig:sup-cache-sort}. Additionally, we assess the performance using metrics such as accuracy, Kendall's Tau, and MAE. Our results show that in the pre-sorting stage (kernel-tile sorting), sorting based on the nearest distance between the \textit{DRK} and the tile achieves the highest sorting accuracy. Sorting based on the most centric approach closely follows in performance. Both methods provide notable improvements compared to cache-sorting alone. Presorting with the nearest distance is also better than the vanilla presorting strategy on \textit{DRK}. For further details and a more in-depth discussion, we refer readers to the StopThePop~\cite{radl2024stopthepop} paper, a pioneering work in this field.



\end{document}